\documentclass{article}
\usepackage{arxiv}

\usepackage[utf8]{inputenc} 
\usepackage[T1]{fontenc}    
\usepackage{hyperref}       
\usepackage{url}            
\usepackage{booktabs}       
\usepackage{amsfonts}       
\usepackage{nicefrac}       
\usepackage{microtype}      
\usepackage{graphicx}
\usepackage{enumerate}
\usepackage{animate}
\usepackage{scalerel}
\usepackage{natbib}
\usepackage[title]{appendix}
\usepackage{epstopdf}
\usepackage[flushleft]{threeparttable}
\usepackage{multirow}

\usepackage{amssymb}
\usepackage{booktabs}
\usepackage{mathtools}
\usepackage{makecell}
\usepackage[inline]{enumitem}
\usepackage[amsmath,thmmarks]{ntheorem}
\usepackage{multicol}
\usepackage{tikz} 
\usetikzlibrary{calc,backgrounds}
\usetikzlibrary{shapes.geometric, arrows, arrows.meta}
\usetikzlibrary{positioning}
\usetikzlibrary{decorations.pathreplacing,calligraphy}
\usetikzlibrary{fadings}
\usepackage{bbm}
\usepackage{xcolor}

\definecolor{vir1}{RGB}{33, 145, 140}
\definecolor{vir2}{RGB}{53, 183, 121}
\definecolor{vir3}{RGB}{181, 222, 43}

\usepackage[caption=false,subrefformat=parens,labelformat=parens]{subfig}
\usepackage{algorithmic}
\usepackage{algorithm}
\ifpdf
  \DeclareGraphicsExtensions{.eps,.pdf,.png,.jpg}
\else
  \DeclareGraphicsExtensions{.eps}
\fi

\theoremstyle{plain}

\theoremstyle{nonumberplain}
\theorembodyfont{\normalfont}
\theoremsymbol{\ensuremath{\square}}

\makeatother

\title{Linked Deep Gaussian Process Emulation for Model Networks}

\author{
  Deyu Ming\thanks{Corresponding author: \texttt{deyu.ming.16@ucl.ac.uk}.} \\
  School of Management\\
  University College London, UK \\
   \And
  Daniel Williamson \\
  Department of Mathematics and Statistics\\
  University of Exeter, UK \\
}

\begin{document}
\maketitle

\begin{abstract}
Modern scientific problems are often multi-disciplinary and require integration of computer models from different disciplines, each with distinct functional complexities, programming environments, and computation times. Linked Gaussian process (LGP) emulation tackles this challenge through a divide-and-conquer strategy that integrates Gaussian process emulators of the individual computer models in a network. However, the required stationarity of the component Gaussian process emulators within the LGP framework limits its applicability in many real-world applications. In this work, we conceptualize a network of computer models as a deep Gaussian process with partial exposure of its hidden layers. We develop a method for inference for these partially exposed deep networks that retains a key strength of the LGP framework, whereby each model can be emulated separately using a DGP and then linked together. We show in both synthetic and empirical examples that our linked deep Gaussian process emulators exhibit significantly better predictive performance than standard LGP emulators in terms of accuracy and uncertainty quantification. They also outperform single DGPs fitted to the network as a whole because they are able to integrate information from the partially exposed hidden layers. Our methods are implemented in an R package \texttt{dgpsi} that is freely available on CRAN.
\end{abstract}

\keywords{surrogate modeling \and multi-physics emulation \and model integration \and network emulation}

\section{Introduction}
\label{sec:intro}
Multi-disciplinary research is at the epicenter of many modern scientific problems, which often require integration of computer models developed in different fields to represent and understand sophisticated real-world engineering, physical or social networks. However, computer models can be expensive to run and networks of computer models can soon become computationally prohibitive if a large number of simulations are required, hindering the analysis of such networks. Statistical emulators or surrogate models are thus needed to accelerate the simulations of the computer model networks, rendering efficient downstream analysis such as sensitivity analysis, calibration, and optimization, particularly when there are only limited computational resources.

Gaussian Process (GP) emulators are statistical models that are widely used in computer model experiments~\citep{jandarov2014emulating,salmanidou2017statistical,donnelly2022gaussian} to emulate computationally expensive simulators because of their flexibility and native uncertainty quantification. Despite their popularity, GP emulators can be inefficient for emulating networks of computer models because they only consider the information contained in the global Inputs/Outputs (I/O) without taking the network structure into account. Linked Gaussian Process (LGP) emulators~\citep{kyzyurova2018coupling,ming2021linked} offer a \emph{divide-and-conquer} approach that allows analytically tractable and flexible surrogate constructions for computer model networks by linking GP emulators of their individual sub-models. By incorporating network structures and internal I/O, LGP emulators can often achieve significantly higher accuracy than GP emulators when the composition of computer models induces non-linear functional forms. However, performance of LGP emulators is largely constrained by the performance of the GP components. 

GP emulators are typically assumed to be stationary, limiting their skill for computer models that exhibit the varying regimes or sharp transitions often seen in real-world applications. Different approaches~\citep{paciorek2003nonstationary,gramacy2008bayesian,montagna2016computer,volodina2020diagnostics} have been proposed to address the stationarity of GP emulators, but one cannot integrate them into the framework of LGP emulation without losing analytical tractability and thus a great deal of efficiency. Deep Gaussian Processes (DGPs), which are extensively explored in the Machine Learning community~\citep{damianou2013deep, wang2016sequential,salimbeni2017doubly,havasi2018inference}, offer a class of models with rich expressiveness and have been shown~\citep{sauer2020active,ming2022deep} to be excellent candidates as emulators of non-stationary computer models. In this paper we exploit two related ideas: First, that an LGP emulator can be thought of as a DGP with full exposure of the latent layers and, second, that a DGP can be thought of as modeling a network of models that are all hidden. These ideas together suggest a generalization to the LGP approaches, modeling a network of computer models as a single DGP with partial exposure of its hidden layers. We will term these emulators Linked Deep Gaussian Process (LDGP) emulators.

Before introducing LDGP emulators in Section~\ref{sec:ldgp}, we first review GP, LGP, and DGP emulators in Section~\ref{sec:gp},~\ref{sec:lgp} and~\ref{sec:dgp} respectively, which naturally give rise to an efficient inference algorithm to build LDGP emulators that we detail in Section~\ref{sec:ldgp}. The performance of LDGP emulators are then demonstrated in a synthetic experiment and an empirical application in Section~\ref{sec:synthetic} and~\ref{sec:empirical} respectively.  

\section{Emulation Frameworks}\label{sec:emulation}
\subsection{Gaussian Process Emulation}
\label{sec:gp}
Let $\mathbf{x}\in\mathbb{R}^{M\times D}$ be an $M$-point design of $D$-dimensional inputs to a computer model with corresponding scalar-valued output $\mathbf{Y}(\mathbf{x})\in\mathbb{R}^{M\times 1}$. In this work, we assume that $\mathbf{Y}(\mathbf{x})$ follows a zero-mean multivariate normal distribution
$
\mathbf{Y}(\mathbf{x})\sim\mathcal{N}(\mathbf{0},\,\sigma^2\mathbf{R}(\mathbf{x})),
$
where $\sigma^2$ is the scale parameter and $\mathbf{R}(\mathbf{x})\in\mathbb{R}^{M\times M}$ is the correlation matrix. The $ij$-th element of $\mathbf{R}(\mathbf{x})$ is specified by $k(\mathbf{x}_{j*},\mathbf{x}_{i*})=
\prod_{d=1}^D k_d(|x_{id}-x_{jd}|)+\eta\prod_{d=1}^D \mathbbm{1}_{\{x_{id}=x_{jd}\}}$, where $k_d(\cdot)$ is a one-dimensional isotropic kernel function that corresponds to the $d$-th input dimension; $\eta$ is the nugget parameter; and $\mathbbm{1}_{\{\cdot\}}$ is the indicator function. Squared exponential and Matérn kernels~\citep{rasmussen2005gaussian} are popular candidates of $k_d(\cdot)$ for GP emulators and often include unknown hyper-parameter $\gamma_d$ to be estimated (through e.g., maximum likelihood) together with $\sigma^2$ and $\eta$, given realizations of $\mathbf{y}=(y_1,\dots,y_M)^\top$ of the computer model output $\mathbf{Y}$. 

Given $\mathbf{x}$ and $\mathbf{y}$, and estimated $\sigma^2$, $\eta$ and $\boldsymbol{\gamma}=(\gamma_1,\dots,\gamma_D)$, the GP emulator of the computer model is defined as the posterior predictive distribution of $Y(\mathbf{x}^*)$ at a new input position $\mathbf{x}^*\in\mathbb{R}^{1\times D}$, which is normal with mean $\mu(\mathbf{x}^*)$ and variance $\sigma^2(\mathbf{x}^*)$ given by:
\begin{equation}
\label{eq:gp}
\mu(\mathbf{x}^*)=\mathbf{r}(\mathbf{x}^*)^\top\mathbf{R}(\mathbf{x})^{-1}\mathbf{y}\quad\mathrm{and}\quad
\sigma^2(\mathbf{x}^*)=\sigma^2\left(1+\eta-\mathbf{r}(\mathbf{x}^*)^\top\mathbf{R}(\mathbf{x})^{-1}\mathbf{r}(\mathbf{x}^*)\right),
\end{equation}
where $\mathbf{r}(\mathbf{x}^*)=[k(\mathbf{x}^*,\mathbf{x}_{1*}),\dots,k(\mathbf{x}^*,\mathbf{x}_{M*})]^\top$.

\subsection{Linked Gaussian Process Emulation}
\label{sec:lgp}
\begin{figure}[!ht]
\centering
\scalebox{0.9}{
\begin{tikzpicture}[shorten >=1pt,->,draw=black!50, node distance=4cm]
    \tikzstyle{every pin edge}=[<-,shorten <=1pt]
    \tikzstyle{neuron}=[circle,fill=black!25,minimum size=35pt,inner sep=0pt]
    \tikzstyle{layer1}=[neuron, fill=vir1];
    \tikzstyle{layer2}=[neuron, fill=vir2];
    \tikzstyle{layer3}=[neuron, fill=vir3];
    \tikzstyle{innerlayer}=[neuron, fill=white];
    \tikzstyle{annot} = [text width=4em, text centered]

    \node[layer1, pin=left:] (l1-0) at (0,0.2) {$\mathcal{GP}_{1,1}$};
    \node[layer1, pin=left:] (l1-1) at (0,-1.25) {$\mathcal{GP}_{1,2}$};
    \node[layer1, pin=left:] (l1-2) at (0,-3.15) {$\mathcal{GP}_{1,P_1}$};
    \node[layer2] (l2-0) at (2.5,0.2) {$\mathcal{GP}_{2,1}$};
    \node[layer2] (l2-1) at (2.5,-1.25) {$\mathcal{GP}_{2,2}$};
    \node[layer2] (l2-2) at (2.5,-3.15) {$\mathcal{GP}_{2,P_2}$};
    \node[innerlayer] (I-0) at (5,0) {\ldots};
    \node[innerlayer] (I-1) at (5,-1.25) {\ldots};
    \node[innerlayer] (I-2) at (5,-2.95) {\ldots};
    \node[layer3,pin={[pin edge={->}]right:}] (ln-0) at (7.5,0.2) {$\mathcal{GP}_{L,1}$};
    \node[layer3, pin={[pin edge={->}]right:}] (ln-1) at (7.5,-1.25) {$\mathcal{GP}_{L,2}$};
    \node[layer3, pin={[pin edge={->}]right:}] (ln-2) at (7.5,-3.15) {$\mathcal{GP}_{L,P_L}$};
\path (l1-1) -- (l1-2) node [black, midway, sloped] {$\dots$};
\path (l2-1) -- (l2-2) node [black, midway, sloped] {$\dots$};
\path (ln-1) -- (ln-2) node [black, midway, sloped] {$\dots$};
\path (I-1) -- (I-2) node [black, midway, sloped] {$\dots$};
\draw[->] (l1-0) -- (l2-0);
\draw[->] (l1-0) -- (l2-1);
\draw[->] (l1-0) -- (l2-2);
\draw[->] (l1-1) -- (l2-0);
\draw[->] (l1-1) -- (l2-1);
\draw[->] (l1-1) -- (l2-2);
\draw[->] (l1-2) -- (l2-0);
\draw[->] (l1-2) -- (l2-1);
\draw[->] (l1-2) -- (l2-2);
\draw[->,path fading=east] (l2-0) -- (I-0);
\draw[->,path fading=east] (l2-0) -- (I-1);
\draw[->,path fading=east] (l2-0) -- (I-2);
\draw[->,path fading=east] (l2-1) -- (I-0);
\draw[->,path fading=east] (l2-1) -- (I-1);
\draw[->,path fading=east] (l2-1) -- (I-2);
\draw[->,path fading=east] (l2-2) -- (I-0);
\draw[->,path fading=east] (l2-2) -- (I-1);
\draw[->,path fading=east] (l2-2) -- (I-2);
\draw[->,path fading=west] (I-0) -- (ln-0);
\draw[->,path fading=west] (I-0) -- (ln-1);
\draw[->,path fading=west] (I-0) -- (ln-2);
\draw[->,path fading=west] (I-1) -- (ln-0);
\draw[->,path fading=west] (I-1) -- (ln-1);
\draw[->,path fading=west] (I-1) -- (ln-2);
\draw[->,path fading=west] (I-2) -- (ln-0);
\draw[->,path fading=west] (I-2) -- (ln-1);
\draw[->,path fading=west] (I-2) -- (ln-2);
\end{tikzpicture}}
\caption{Hierarchy of an LGP emulator.}
\label{fig:lgpmodel}
\end{figure}
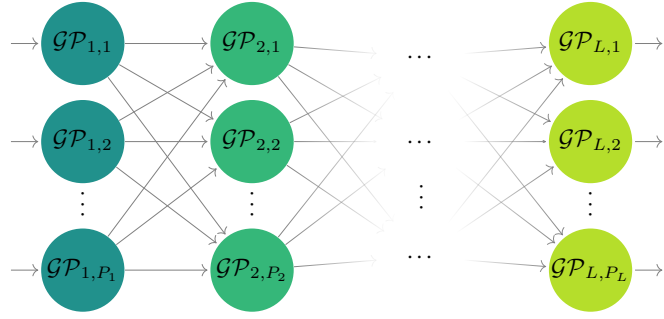

LGP emulators emulate feed-forward networks of computer models by linking GP emulators of individual computer models. Consider a $L$-layered network of computer models where each output dimension of a computer model is emulated by a GP emulator $\mathcal{GP}_{l,p}$ for $p=1,\dots,P_l$ and $l=1,\dots,L$, where $P_l$ is the total number of output dimensions of computer models in layer $l$ of the network, see Figure~\ref{fig:lgpmodel}. 

Let $\mathbf{x}_{l,p}\in\mathbb{R}^{M_{l,p}\times D_{l,p}}$ be $M_{l,p}$ sets of $D_{l,p}$-dimensional input and $\mathbf{Y}_{l,p}\in\mathbb{R}^{M_{l,p}\times 1}$ be the corresponding one-dimensional output of $\mathcal{GP}_{l,p}$, where $D_{l,p}\leq P_{l-1}$ for $l=2,\dots,L$. Given realizations $\mathbf{y}_{l,p}$ of $\mathbf{Y}_{l,p}$ for all $p=1,\dots,P_l$ and $l=1,\dots,L$, the analytically intractable posterior predictive distribution of output $\mathbf{Y}(\mathbf{x}^*)=[{Y}_{L,1}(\mathbf{x}^*),\dots,{Y}_{L,P_L}(\mathbf{x}^*)]$ at a new input position $\mathbf{x}^*=\{\mathbf{x}^*_{1,1},\dots,\mathbf{x}^*_{1,P_1}\}$, where $\mathbf{x}^*_{1,p}\in\mathbb{R}^{1\times D_{1,p}}$ for $p=1,\dots,P_1$, can be sufficiently approximated by a normal distribution with closed form mean and variance, under some mild conditions~\citep{ming2021linked}. This normal approximation defines the LGP emulator, and one can obtain the mean $\boldsymbol{\mu}_{1\rightarrow L}(\mathbf{x}^*)=[{\mu}_{1\rightarrow L,1}(\mathbf{x}^*),\dots,{\mu}_{1\rightarrow L,P_L}(\mathbf{x}^*)]$ and variance $\boldsymbol{\sigma}^2_{1\rightarrow L}(\mathbf{x}^*)=[{\sigma}_{1\rightarrow L,1}^2(\mathbf{x}^*),\dots,{\sigma}_{1\rightarrow L,P_L}^2(\mathbf{x}^*)]$ by iterating the following formulae:
\begin{align}
\label{eq:linkgp_mean}
{\mu}_{1\rightarrow l,p}(\mathbf{x}^*)=&\mathbf{I}_{l,p}(\mathbf{x}^*)^\top\mathbf{R}_{l,p}^{-1}\mathbf{y}_{l,p},\\
\label{eq:linkgp_var}
{\sigma}_{1\rightarrow l,p}^2(\mathbf{x}^*)=&\mathbf{y}_{l,p}^\top\mathbf{R}_{l,p}^{-1}\mathbf{J}_{l,p}(\mathbf{x}^*)\mathbf{R}_{l,p}^{-1}\mathbf{y}_{l,p}-\left(\mathbf{I}_{l,p}(\mathbf{x}^*)^\top\mathbf{R}_{l,p}^{-1}\mathbf{y}_{l,p}\right)^2+\sigma^2_{l,p}\left(1+\eta_{l,p}-\mathrm{tr}\left\{\mathbf{R}_{l,p}^{-1}\mathbf{J}_{l,p}(\mathbf{x}^*)\right\}\right)
\end{align}
for $l=2,\dots,L$ and $p=1,\dots,P_l$, with the $i$-th element of $\mathbf{I}_{l,p}(\mathbf{x}^*)\in\mathbb{R}^{M_{l,p}\times 1}$ given by 
    $$
    \prod_{q\in\mathbb{S}^{D_{l,p}}_{l-1}}\xi_{l,p}\left({\mu}_{1\rightarrow(l-1),q}(\mathbf{x}^*),\,{\sigma}^{2}_{1\rightarrow(l-1),q}(\mathbf{x}^*),\,(\mathbf{x}_{l,p})_{iq}\right),
    $$
and the $ij$-th element of $\mathbf{J}_{l,p}(\mathbf{x}^*)\in\mathbb{R}^{M_{l,p}\times M_{l,p}}$ given by 
    $$
   \prod_{q\in\mathbb{S}^{D_{l,p}}_{l-1}}\zeta_{l,p}({\mu}_{1\rightarrow(l-1),q}(\mathbf{x}^*),\,{\sigma}^{2}_{1\rightarrow(l-1),q}(\mathbf{x}^*),\,(\mathbf{x}_{l,p})_{iq},\,(\mathbf{x}_{l,p})_{jq}),
    $$
where ${\mu}_{1\rightarrow 1,q}(\mathbf{x}^*)$ and ${\sigma}^{2}_{1\rightarrow 1,q}(\mathbf{x}^*)$ are the mean and variance of the GP emulator, $\mathcal{GP}_{1,q}$, for $q=1,\dots,P_1$; $\mathbb{S}^{D_{l,p}}_{l-1}$ is a subset of $\mathbb{S}_{l-1}=\{1,\dots,P_{l-1}\}$ that contains the $D_{l,p}$ indices corresponding to the GP emulators in layer $l-1$ whose outputs are inputs to $\mathcal{GP}_{l,p}$; and $\xi_{l,p}(\cdot,\cdot,\cdot)$ and $\zeta_{l,p}(\cdot,\cdot,\cdot,\cdot)$ are analytically tractable functions with their closed form expressions given in~\citet[Appendix A-D]{ming2021linked}, if the kernel functions of $\mathcal{GP}_{l,p}$ are squared exponentials or Matérns. Given that the unknown model parameters in $\mathcal{GP}_{l,p}$ are estimated (e.g., via maximum likelihood) for all $p=1,\dots,P_l$ and $l=1,\dots,L$, the LGP emulator can be constructed by invoking~\eqref{eq:linkgp_mean} and~\eqref{eq:linkgp_var}, as summarized in Algorithm~\ref{alg:lgp}. 

\begin{algorithm}[htbp]
\caption{Construction of an LGP emulator with the hierarchy in Figure~\ref{fig:lgpmodel}}
\label{alg:lgp}
\begin{algorithmic}[1]
\REQUIRE{\begin{enumerate*}[label=(\roman*)]
  \item Realizations $\mathbf{x}_{l,p}$ and $\mathbf{y}_{l,p}$ for all $p=1,\dots,P_l$ and $l=1,\dots,L$;
  \item A new input position $\mathbf{x}^*=\{\mathbf{x}^*_{1,1},\dots,\mathbf{x}^*_{1,P_1}\}$.
\end{enumerate*}}
\ENSURE{Mean and variance of $\mathbf{Y}(\mathbf{x}^*)=[{Y}_{L,1}(\mathbf{x}^*),\dots,{Y}_{L,P_L}(\mathbf{x}^*)]$.}
\STATE{Compute the mean ${\mu}_{1\rightarrow 1,q}(\mathbf{x}^*)$ and variance  ${\sigma}^{2}_{1\rightarrow 1,q}(\mathbf{x}^*)$ of the first-layer GP emulator $\mathcal{GP}_{1,q}$ by using~\eqref{eq:gp}, for all $p=1,\dots,P_1$.}
\STATE{Compute the mean $\boldsymbol{\mu}_{1\rightarrow L}(\mathbf{x}^*)$ and variance $\boldsymbol{\sigma}^2_{1\rightarrow L}(\mathbf{x}^*)$ of $\mathbf{Y}(\mathbf{x}^*)$ by iterating~\eqref{eq:linkgp_mean} and~\eqref{eq:linkgp_var}.}
\end{algorithmic} 
\end{algorithm}

\subsection{Deep Gaussian Process Emulation}
\label{sec:dgp}
DGP emulators can be used to emulate computer models that exhibit non-stationary functional behaviors. A DGP emulator has the same model hierarchy as the one shown in Figure~\ref{fig:lgpmodel} for an LGP emulator, except that the internal I/O are latent. Although there are different inference approaches~\citep{salimbeni2017doubly,sauer2020active} that can be used to build DGP emulators, the structural similarity between DGP and LGP emulators renders a simple while efficient method, called Stochastic Imputation (SI), that provides a good balance between the computation and accuracy of DGP emulator constructions~\citep{ming2022deep}. The core idea of SI is to convert a DGP emulator into a set of LGP emulators, each of which represents a realization of the DGP emulator with its latent I/O exposed (i.e., imputed). As a result, one can obtain the DGP emulator by constructing and aggregating LGP emulators. 

Let $\mathbf{x}\in\mathbb{R}^{M\times P_1}$ and $\mathbf{Y}\in\mathbb{R}^{M\times P_L}$ be $M$ sets of $P_1$-dimensional input and $P_L$-dimensional output of a computer model, and $\mathbf{W}_{l,p}\in\mathbb{R}^{M\times1}$ be the latent output of $\mathcal{GP}_{l,p}$ in a DGP emulator hierarchy given in Figure~\ref{fig:lgpmodel}, where $p=1,\dots,P_l$ and $l=1,\dots,L-1$. Given realizations $\mathbf{y}$ of $\mathbf{Y}$, one can obtain point estimates of unknown model parameters in $\mathcal{GP}_{l,p}$ for all $p=1,\dots,P_l$ and $l=1,\dots,L$, using the Stochastic Expectation Maximization (SEM) algorithm~\citep{ming2022deep}. With the estimated model parameters, the DGP emulator, which gives the approximate
posterior predictive mean and variance of $\mathbf{Y}(\mathbf{x}^*)=[Y_1(\mathbf{x}^*),\dots,Y_{P_L}(\mathbf{x}^*)]$ at a new input position $\mathbf{x}^*$, can be constructed by following the steps in Algorithm~\ref{alg:dgp}.

\begin{algorithm}[htbp]
\caption{Construction of a DGP emulator with the hierarchy in Figure~\ref{fig:lgpmodel}}
\label{alg:dgp}
\begin{algorithmic}[1]
\REQUIRE{\begin{enumerate*}[label=(\roman*)]
  \item Realizations $\mathbf{x}$ and $\mathbf{y}$;
  \item A new input position $\mathbf{x}^*$;
  \item The number of imputations $N$.
\end{enumerate*}}
\ENSURE{Mean and variance of $\mathbf{Y}(\mathbf{x}^*)$.}
\FOR{$i=1,\dots,N$}
\FOR{\label{alg:imp_start}$l=1,\dots,L-1$}
\STATE{\label{alg:one_ess}Given $\mathbf{x}$ and $\mathbf{y}$, draw an imputation $\{\mathbf{w}^{(i)}_{l,p}\}_{p=1,\dots,P_l}$ of the latent output $\{\mathbf{W}_{l,p}\}_{p=1,\dots,P_l}$ via an Elliptical Slice Sampling~\citep{nishihara2014parallel} update, see Appendix~\ref{app:ess}.}
\ENDFOR\label{alg:imp_end}
\STATE{Construct the LGP emulator $\mathcal{LGP}_i$ with the mean $\boldsymbol{\mu}^{(i)}_{1\rightarrow L}(\mathbf{x}^*)$ and variance ${\boldsymbol{\sigma}^2}^{(i)}_{1\rightarrow L}(\mathbf{x}^*)$ by Algorithm~\ref{alg:lgp}, given $\mathbf{x}$, $\mathbf{y}$, and $\{\mathbf{w}^{(i)}_{l,p}\}_{p=1,\dots,P_l,\,l=1,\dots,L-1}$.}
\ENDFOR
\STATE{Compute the mean $\boldsymbol{\mu}(\mathbf{x}^*)$ and variance $\boldsymbol{\sigma}^2(\mathbf{x}^*)$ of $\mathbf{Y}(\mathbf{x}^*)$ by
\begin{align*}
 \boldsymbol{\mu}(\mathbf{x}^*)&=\frac{1}{N}\sum_{i=1}^N\boldsymbol{\mu}^{(i)}_{1\rightarrow L}(\mathbf{x}^*),\\
 \boldsymbol{\sigma}^2(\mathbf{x}^*)&=\frac{1}{N}\sum_{i=1}^N\left([\boldsymbol{\mu}^{(i)}_{1\rightarrow L}(\mathbf{x}^*)]^2+{\boldsymbol{\sigma}^2}^{(i)}_{1\rightarrow L}(\mathbf{x}^*)\right)-\boldsymbol{\mu}(\mathbf{x}^*)^2.
\end{align*}
}
\end{algorithmic} 
\end{algorithm}

\section{Linked Deep Gaussian Process Emulation}
\label{sec:ldgp}
LGP emulation provides a analytically tractable way to construct emulators for computer model networks by exploiting network structures and the computational heterogeneity among individual sub-models. However, nonstationarity of any of the underlying sub-models severely impacts the performance of LGP emulators. To address this limitation, a natural extension is to replace GP emulators in an LGP structure with DGP emulators, leading to an LDGP emulator.

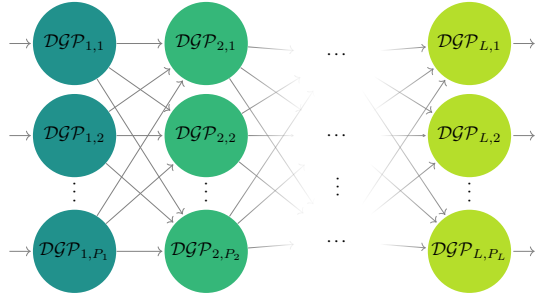
\begin{figure}[!ht]
\centering
\scalebox{0.7}{
\begin{tikzpicture}[shorten >=1pt,->,draw=black!50, node distance=4cm]
    \tikzstyle{every pin edge}=[<-,shorten <=1pt]
    \tikzstyle{neuron}=[circle,fill=black!25,minimum size=45pt,inner sep=0pt]
    \tikzstyle{layer1}=[neuron, fill=vir1];
    \tikzstyle{layer2}=[neuron, fill=vir2];
    \tikzstyle{layer3}=[neuron, fill=vir3];
    \tikzstyle{innerlayer}=[neuron, fill=white];
    \tikzstyle{annot} = [text width=4em, text centered]

    \node[layer1, pin=left:] (l1-0) at (0,0.2) {$\mathcal{DGP}_{1,1}$};
    \node[layer1, pin=left:] (l1-1) at (0,-1.55) {$\mathcal{DGP}_{1,2}$};
    \node[layer1, pin=left:] (l1-2) at (0,-3.75) {$\mathcal{DGP}_{1,P_1}$};
    \node[layer2] (l2-0) at (2.5,0.2) {$\mathcal{DGP}_{2,1}$};
    \node[layer2] (l2-1) at (2.5,-1.55) {$\mathcal{DGP}_{2,2}$};
    \node[layer2] (l2-2) at (2.5,-3.75) {$\mathcal{DGP}_{2,P_2}$};
    \node[innerlayer] (I-0) at (5,0) {\ldots};
    \node[innerlayer] (I-1) at (5,-1.55) {\ldots};
    \node[innerlayer] (I-2) at (5,-3.55) {\ldots};
    \node[layer3,pin={[pin edge={->}]right:}] (ln-0) at (7.5,0.2) {$\mathcal{DGP}_{L,1}$};
    \node[layer3, pin={[pin edge={->}]right:}] (ln-1) at (7.5,-1.55) {$\mathcal{DGP}_{L,2}$};
    \node[layer3, pin={[pin edge={->}]right:}] (ln-2) at (7.5,-3.75) {$\mathcal{DGP}_{L,P_L}$};
\path (l1-1) -- (l1-2) node [black, midway, sloped] {$\dots$};
\path (l2-1) -- (l2-2) node [black, midway, sloped] {$\dots$};
\path (ln-1) -- (ln-2) node [black, midway, sloped] {$\dots$};
\path (I-1) -- (I-2) node [black, midway, sloped] {$\dots$};
\draw[->] (l1-0) -- (l2-0);
\draw[->] (l1-0) -- (l2-1);
\draw[->] (l1-0) -- (l2-2);
\draw[->] (l1-1) -- (l2-0);
\draw[->] (l1-1) -- (l2-1);
\draw[->] (l1-1) -- (l2-2);
\draw[->] (l1-2) -- (l2-0);
\draw[->] (l1-2) -- (l2-1);
\draw[->] (l1-2) -- (l2-2);
\draw[->,path fading=east] (l2-0) -- (I-0);
\draw[->,path fading=east] (l2-0) -- (I-1);
\draw[->,path fading=east] (l2-0) -- (I-2);
\draw[->,path fading=east] (l2-1) -- (I-0);
\draw[->,path fading=east] (l2-1) -- (I-1);
\draw[->,path fading=east] (l2-1) -- (I-2);
\draw[->,path fading=east] (l2-2) -- (I-0);
\draw[->,path fading=east] (l2-2) -- (I-1);
\draw[->,path fading=east] (l2-2) -- (I-2);
\draw[->,path fading=west] (I-0) -- (ln-0);
\draw[->,path fading=west] (I-0) -- (ln-1);
\draw[->,path fading=west] (I-0) -- (ln-2);
\draw[->,path fading=west] (I-1) -- (ln-0);
\draw[->,path fading=west] (I-1) -- (ln-1);
\draw[->,path fading=west] (I-1) -- (ln-2);
\draw[->,path fading=west] (I-2) -- (ln-0);
\draw[->,path fading=west] (I-2) -- (ln-1);
\draw[->,path fading=west] (I-2) -- (ln-2);
\end{tikzpicture}}
\caption{Hierarchy of an LDGP emulator.}
\label{fig:ldgp}
\end{figure}

The LDGP emulator creates a hyper-deep structural hierarchy (shown in Figure~\ref{fig:ldgp}), and DGP components in this hierarchy render the inference challenging because one cannot utilize Algorithm~\ref{alg:lgp} to link DGP emulators analytically. Noting that the hyper-deep structural hierarchy in Figure~\ref{fig:ldgp} itself represents a deep Gaussian process with partial exposure of its hidden layers, enabling a SI approach for LDGP inferences.

Let $\mathbf{x}_{l,p}\in\mathbb{R}^{M_{l,p}\times D_{l,p}}$ be $M_{l,p}$ sets of $D_{l,p}$-dimensional input and $\mathbf{Y}_{l,p}\in\mathbb{R}^{M_{l,p}\times Q_{l,p}}$ be the corresponding $Q_{l,p}$-dimensional output of the DGP component $\mathcal{DGP}_{l,p}$ in an LDGP hierarchy, where $D_{l,p}\leq \sum_{p=1}^{P_{l-1}}Q_{l-1,p}$ for $l=2,\dots,L$. Then, given $\mathbf{x}_{l,p}$ and realizations $\mathbf{y}_{l,p}$ of $\mathbf{Y}_{l,p}$ for all $p=1,\dots,P_l$ and $l=1,\dots,L$, we can convert the LDGP network to an LGP network by imputing the latent variables $\{\mathcal{W}_{l,p}\}_{p=1,\dots,P_l,\,l=1,\dots,L}$ in the LDGP hierarchy (i.e., the hidden layers in its DGP components) through:
\begin{equation*}
p(\{\mathcal{W}_{l,p}\}_{p=1,\dots,P_l,\,l=1,\dots,L}|\{\mathbf{x}_{l,p}\}_{p=1,\dots,P_l,\,l=1,\dots,L},\,\{\mathbf{y}_{l,p}\}_{p=1,\dots,P_l,\,l=1,\dots,L}),
\end{equation*}
which amounts to independent imputations of latent layers in individual DGP emulators contained in the LDGP network, since we have
\begin{equation*}
p(\{\mathcal{W}_{l,p}\}_{p=1,\dots,P_l,\,l=1,\dots,L}|\{\mathbf{x}_{l,p}\}_{p=1,\dots,P_l,\,l=1,\dots,L},\,\{\mathbf{y}_{l,p}\}_{p=1,\dots,P_l,\,l=1,\dots,L})=\prod_{l,p}p(\mathcal{W}_{l,p}|\mathbf{x}_{l,p},\mathbf{y}_{l,p}).
\end{equation*}
As a result, given that the unknown model parameters in $\mathcal{DGP}_{l,p}$ are estimated via SEM for all $p=1,\dots,P_l$ and $l=1,\dots,L$, the LDGP emulator can be constructed via~Algorithm~\ref{alg:ldgp}.

\begin{algorithm}[!ht]
\caption{Construction of an LDGP emulator with the hierarchy in Figure~\ref{fig:ldgp}}
\label{alg:ldgp}
\begin{algorithmic}[1]
\REQUIRE{\begin{enumerate*}[label=(\roman*)]
  \item Realizations $\mathbf{x}_{l,p}$ and $\mathbf{y}_{l,p}$ for all $p=1,\dots,P_l$ and $l=1,\dots,L$;
  \item A new input position $\mathbf{x}^*=\{\mathbf{x}^*_{1,1},\dots,\mathbf{x}^*_{1,P_1}\}$;
  \item The number of imputations $N$.
\end{enumerate*}}
\ENSURE{Mean and variance of $\mathbf{Y}_{L,p}(\mathbf{x}^*)$ for all $p=1,\dots,P_L$.}
\FOR{$i=1,\dots,N$}
\STATE{Given $\mathbf{x}_{l,p}$ and $\mathbf{y}_{l,p}$, generate an imputation $\mathcal{W}^{(i)}_{l,p}$ of $\mathcal{W}_{l,p}$ for all $p=1,\dots,P_l$ and $l=1,\dots,L$ following Step~\ref{alg:imp_start} to~\ref{alg:imp_end} in Algorithm~\ref{alg:dgp}.}
\STATE{Construct the LGP emulator $\mathcal{LGP}_i$ with its mean $\boldsymbol{\mu}^{(i)}_{1\rightarrow L}(\mathbf{x}^*)$ and variance ${\boldsymbol{\sigma}^2}^{(i)}_{1\rightarrow L}(\mathbf{x}^*)$ computed via Algorithm~\ref{alg:lgp}, given $\mathbf{x}_{l,p}$, $\mathbf{y}_{l,p}$, and $\mathcal{W}^{(i)}_{l,p}$ for all $p=1,\dots,P_l$ and $l=1,\dots,L$.}
\ENDFOR
\STATE{Compute the mean $\boldsymbol{\mu}_{1\rightarrow L,p}(\mathbf{x}^*)$ and variance $\boldsymbol{\sigma}^2_{1\rightarrow L,p}(\mathbf{x}^*)$ of $\mathbf{Y}_{L,p}(\mathbf{x}^*)$ by
\begin{align*}
 \boldsymbol{\mu}_{1\rightarrow L,p}(\mathbf{x}^*)&=\frac{1}{N}\sum_{i=1}^N\boldsymbol{\mu}^{(i)}_{1\rightarrow L,p}(\mathbf{x}^*),\\
 \boldsymbol{\sigma}^2_{1\rightarrow L,p}(\mathbf{x}^*)&=\frac{1}{N}\sum_{i=1}^N\left([\boldsymbol{\mu}^{(i)}_{1\rightarrow L,p}(\mathbf{x}^*)]^2+{\boldsymbol{\sigma}^2}^{(i)}_{1\rightarrow L,p}(\mathbf{x}^*)\right)-\boldsymbol{\mu}_{1\rightarrow L,p}(\mathbf{x}^*)^2
\end{align*}
for all $p=1,\dots,P_L$.
}
\end{algorithmic} 
\end{algorithm}

Note that the treatment of a model network as single DGP offers two particular advantages within our framework. First, it suggests a natural inferential approach to the network problem via a slight augmentation to the method (see Algorithm~\ref{alg:dgp}) in~\citet{ming2022deep}, where some of the hidden layers need not be imputed as they are observed. Second, it amounts to modeling the individual components of the network via DGPs without having to run the entire network to obtain each training point, enabling us to obtain more runs and better component emulators for sub-models that are relatively inexpensive to run. We now present some examples and say more about these properties in the discussion.

\section{A One-Dimensional Synthetic Experiment}
\label{sec:synthetic}
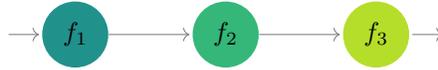
\begin{figure}[!ht]
\centering 
\begin{tikzpicture}[shorten >=1pt,->,draw=black!50, node distance=4cm]
    \tikzstyle{every pin edge}=[<-,shorten <=1pt]
    \tikzstyle{neuron}=[circle,fill=vir2,minimum size=25pt,inner sep=0pt]
    \tikzstyle{layer1}=[neuron, fill=vir1];
    \tikzstyle{layer2}=[neuron, fill=vir2];
    \tikzstyle{layer3}=[neuron, fill=vir3];
    \tikzstyle{annot} = [text width=2.5em, text centered]
    \node[layer1,pin={[pin edge={<-}]left:}] (K-1) at (0, 0) {$f_1$};
    \node[layer2] (K-2) at (2, 0) {$f_2$};
    \node[layer3,pin={[pin edge={->}]right:}] (K-3) at (4, 0) {$f_3$};
    \path[draw] (K-1) -- (K-2);
    \path[draw] (K-2) -- (K-3);
\end{tikzpicture}
\caption{The hierarchy of the synthetic network in Section~\ref{sec:synthetic} that consists of three sequentially connected 1-D computer models $f_1$, $f_2$, and $f_3$.}
\label{fig:1d_example}
\end{figure}

Consider a synthetic network (shown in Figure~\ref{fig:1d_example}) that consists of three 1-D models 
\begin{align}
\label{eq:f1}
f_1(x)&=\frac{\sin(7.5x)+1}{2},\\
\label{eq:f2}
f_2(x)&=\frac{1}{3}\sin(4x-4)+\frac{2}{3}\exp\{-120(2x-1)^2\}+\frac{1}{3},\\
\label{eq:f3}
f_3(x)&=-\frac{5}{6}(\sin(40(0.4x-0.85)^4)\cos(x-2.375)+0.2x+0.55)+1
\end{align}
that are connected sequentially, where $x\in[0,1]$ for all models. We compare a Composite GP (CGP) emulator, a Composite DGP (CDGP) emulator, and both LGP and LDGP emulators using the R package \texttt{dgpsi} with the squared exponential kernel. For the LDGP emulator, we choose 2-layered DGPs for the 3 individual computer models in the chain. We train the CGP emulator, CDGP emulator, and all individual GP and DGP emulators that comprise the LGP and LDGP emulators adaptively. Specifically, we choose 5 data points (whose input positions are uniformly spaced over $[0,1]$) as the initial designs for all emulators, then refine initial fits by adding 15 design points sequentially to the initial design based on the MICE criterion~\citep{beck2016sequential}. 

\begin{figure}[htbp]
\centering 
\subfloat[CGP Emulator]{\label{fig:gp}\includegraphics[width=0.2475\linewidth]{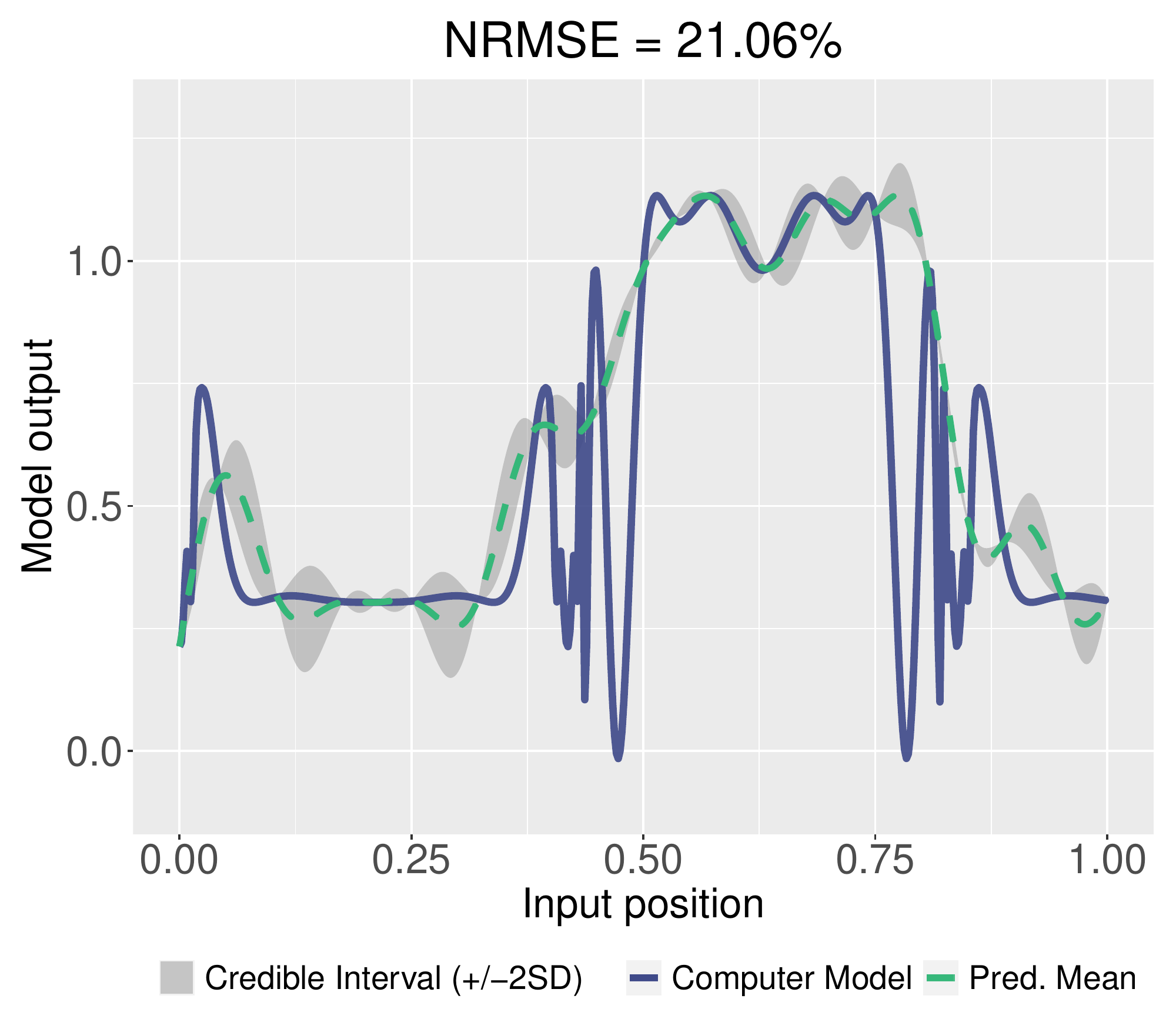}}
\subfloat[CDGP Emulator]{\label{fig:dgp}\includegraphics[width=0.2475\linewidth]{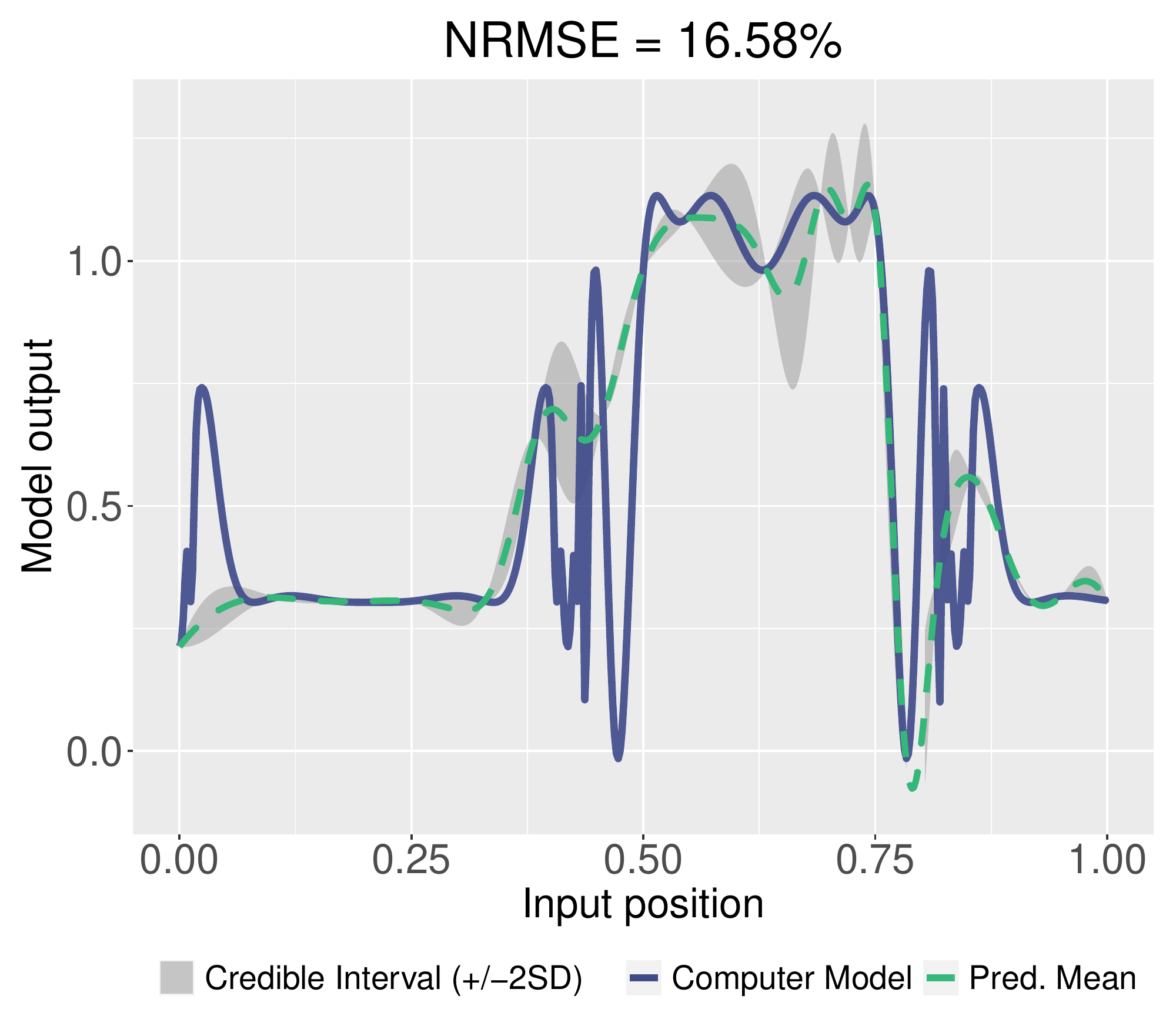}}
\subfloat[LGP Emulator]{\label{fig:lgp}\includegraphics[width=0.2475\linewidth]{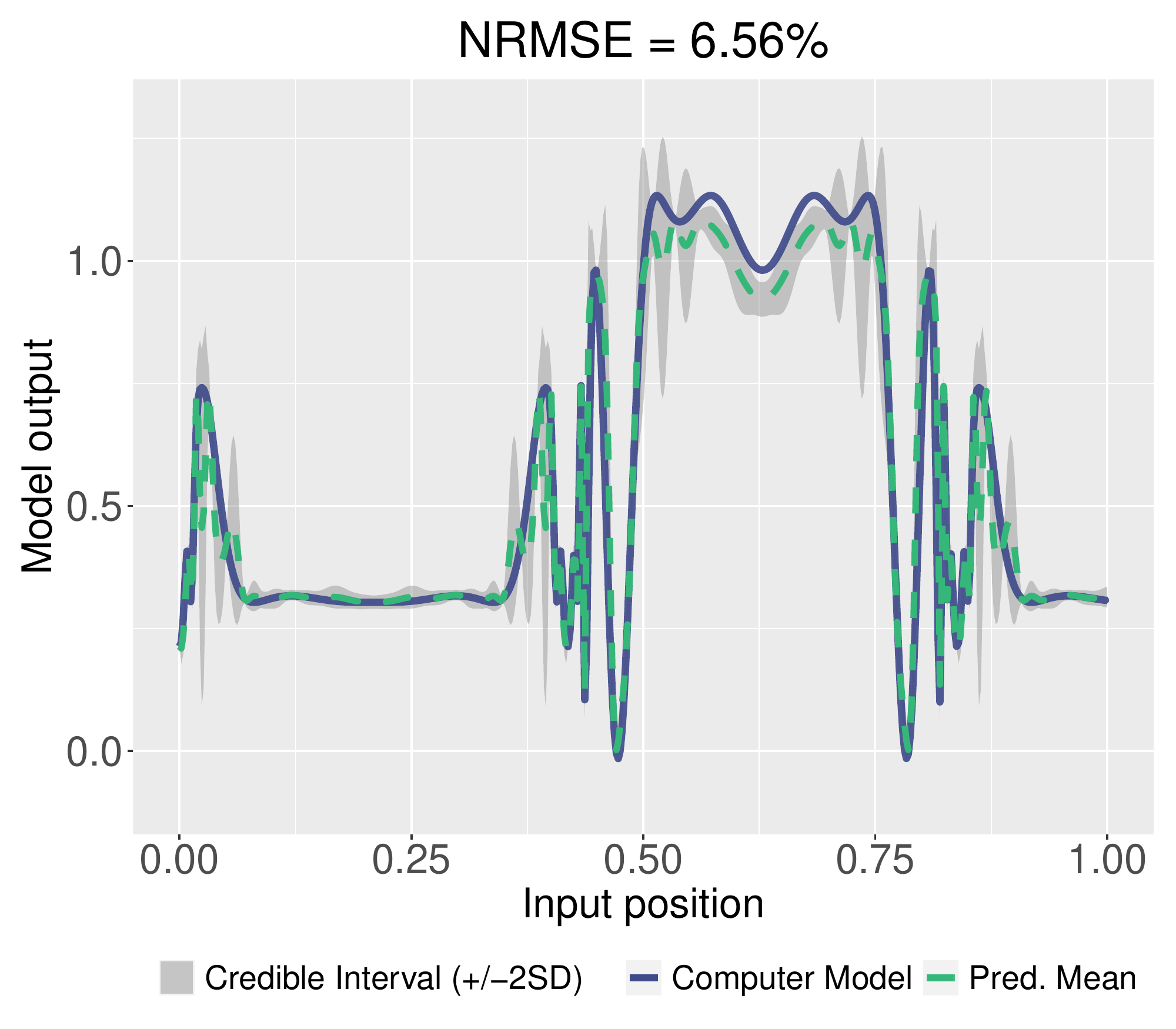}}
\subfloat[LDGP Emulator]{\label{fig:ldgp1}\includegraphics[width=0.2475\linewidth]{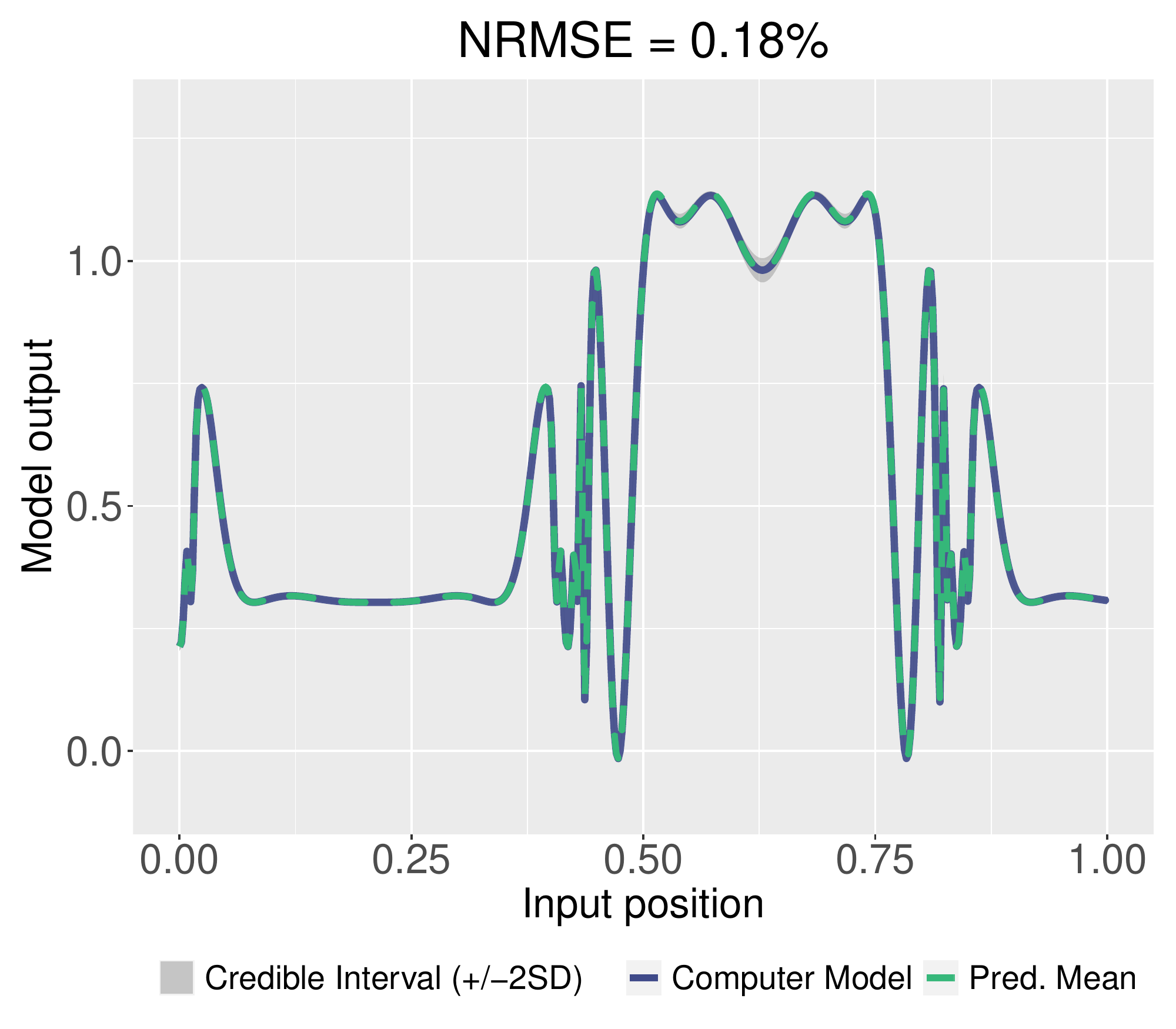}}\\
\subfloat[GP Emulator of $f_1$]{\label{fig:gp1}\includegraphics[width=0.33\linewidth]{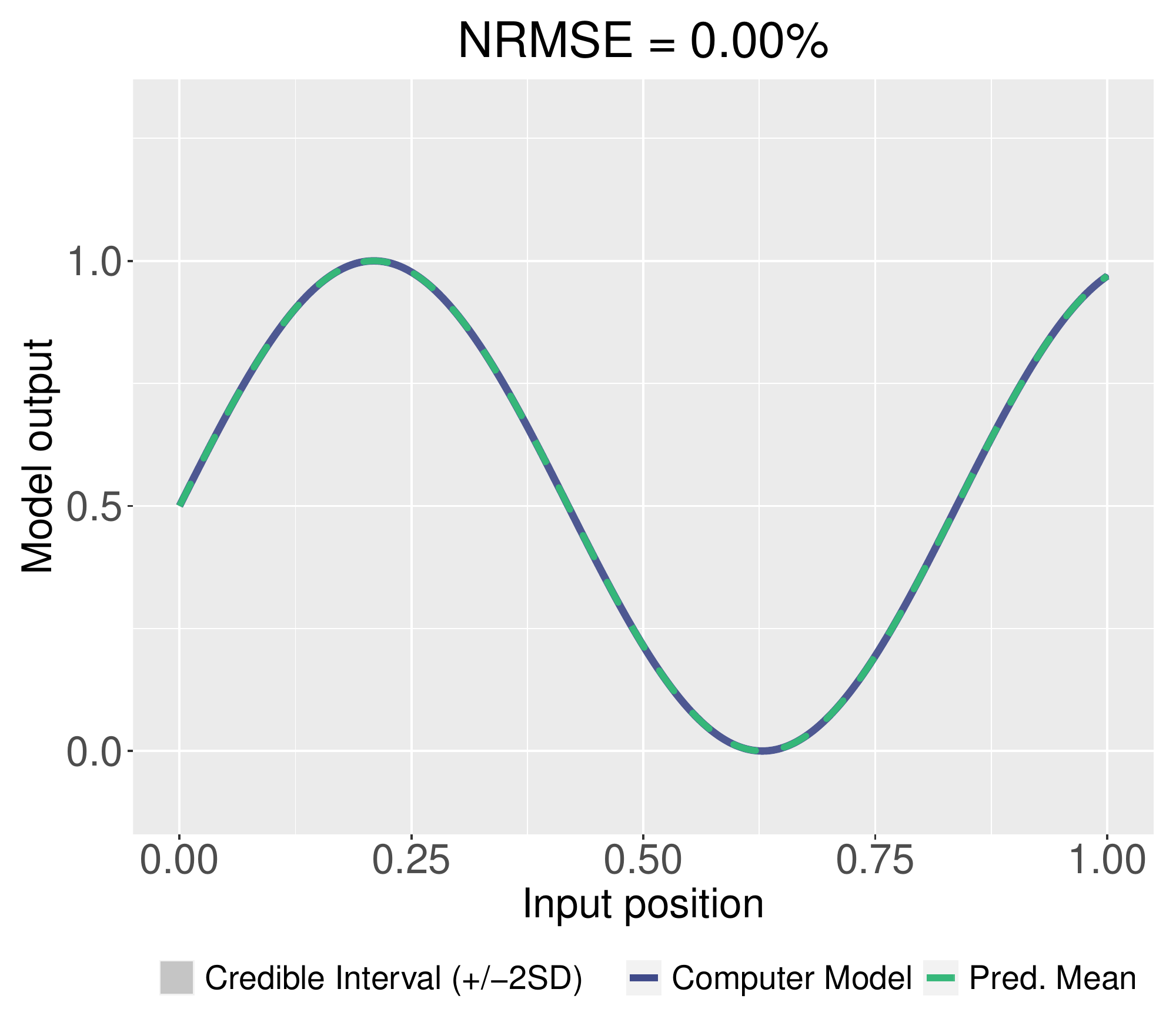}}
\subfloat[GP Emulator of $f_2$]{\label{fig:gp2}\includegraphics[width=0.33\linewidth]{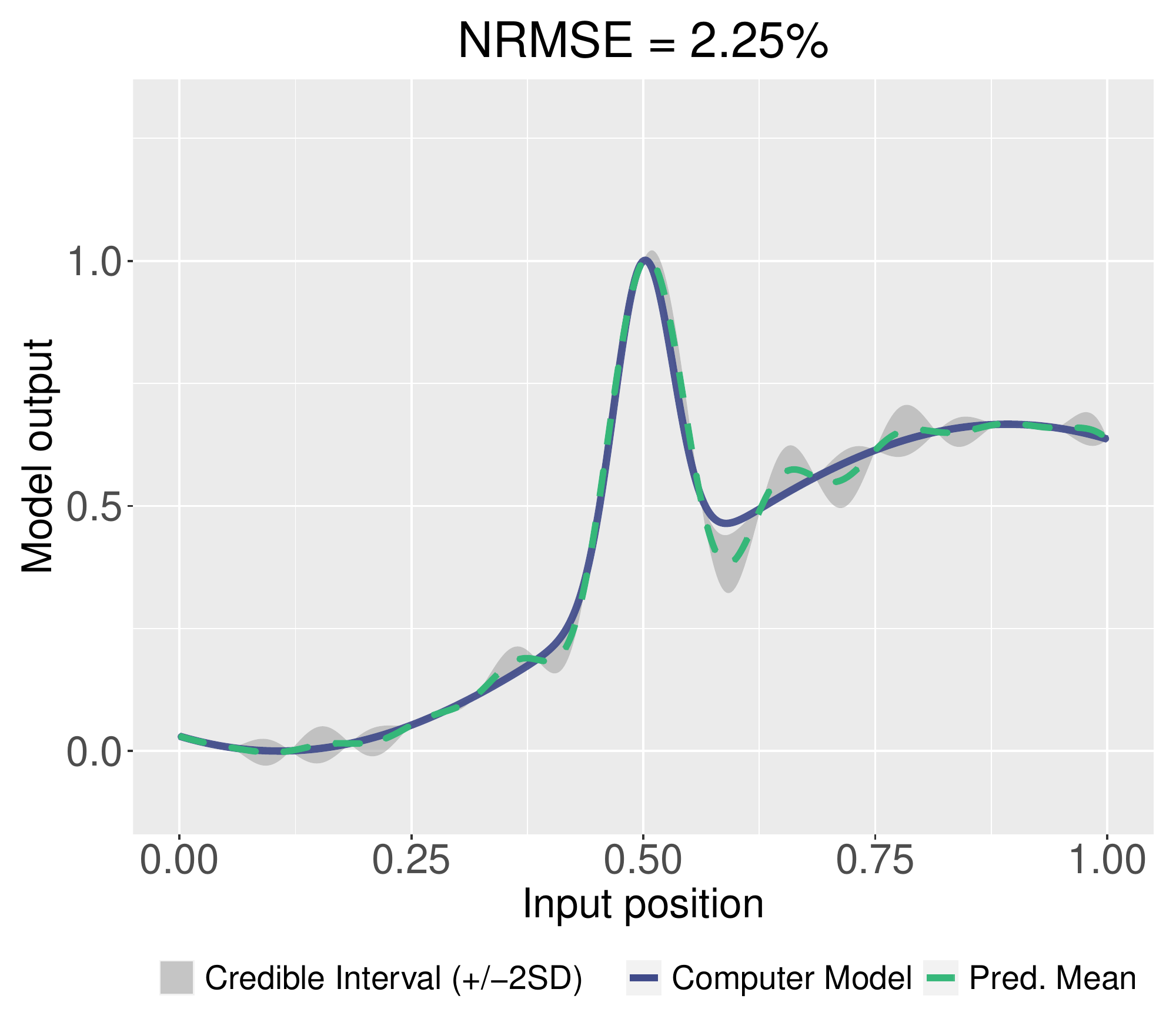}}
\subfloat[GP Emulator of $f_3$]{\label{fig:gp3}\includegraphics[width=0.33\linewidth]{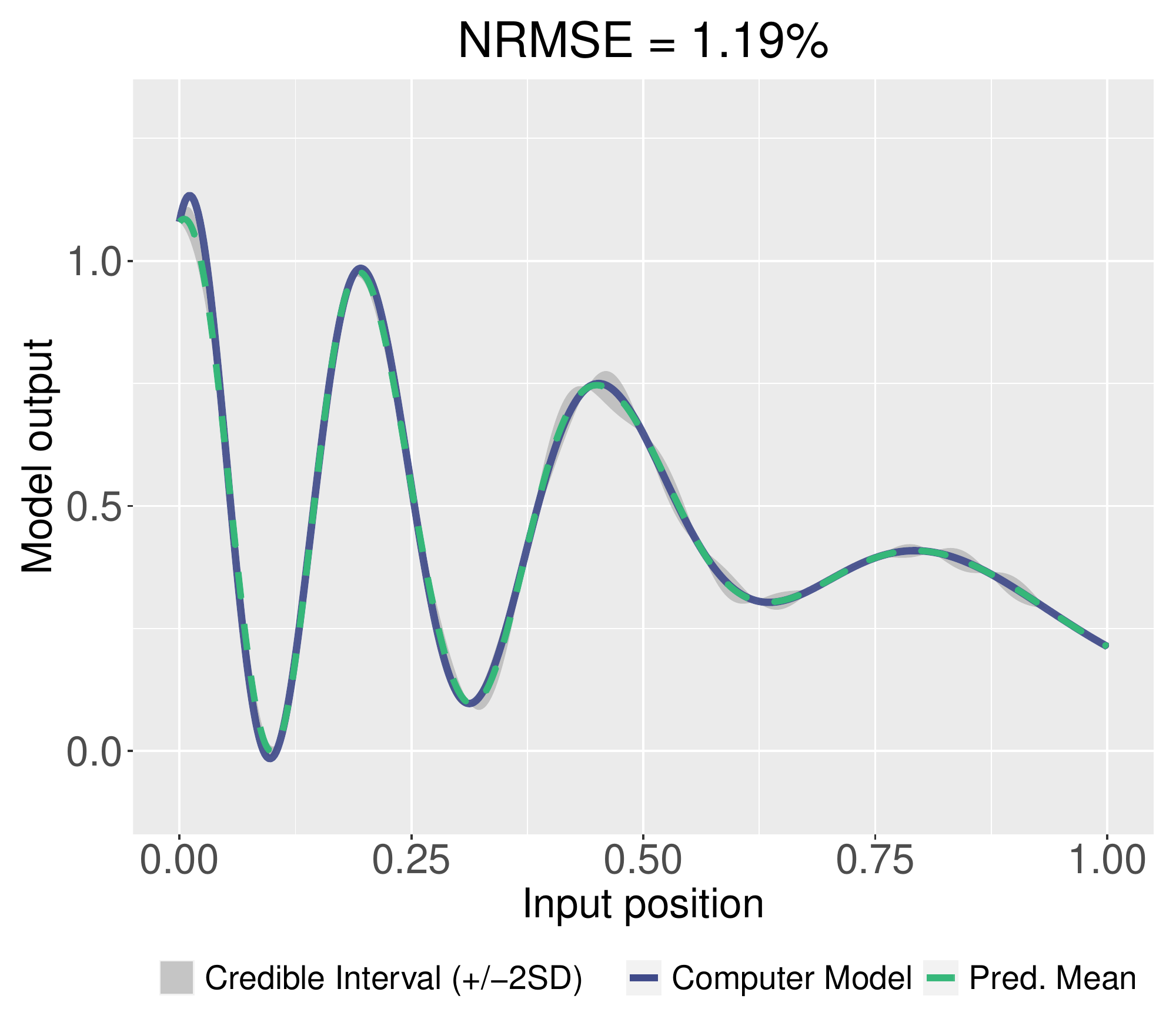}}\\
\subfloat[DGP Emulator of $f_1$]{\label{fig:dgp1}\includegraphics[width=0.33\linewidth]{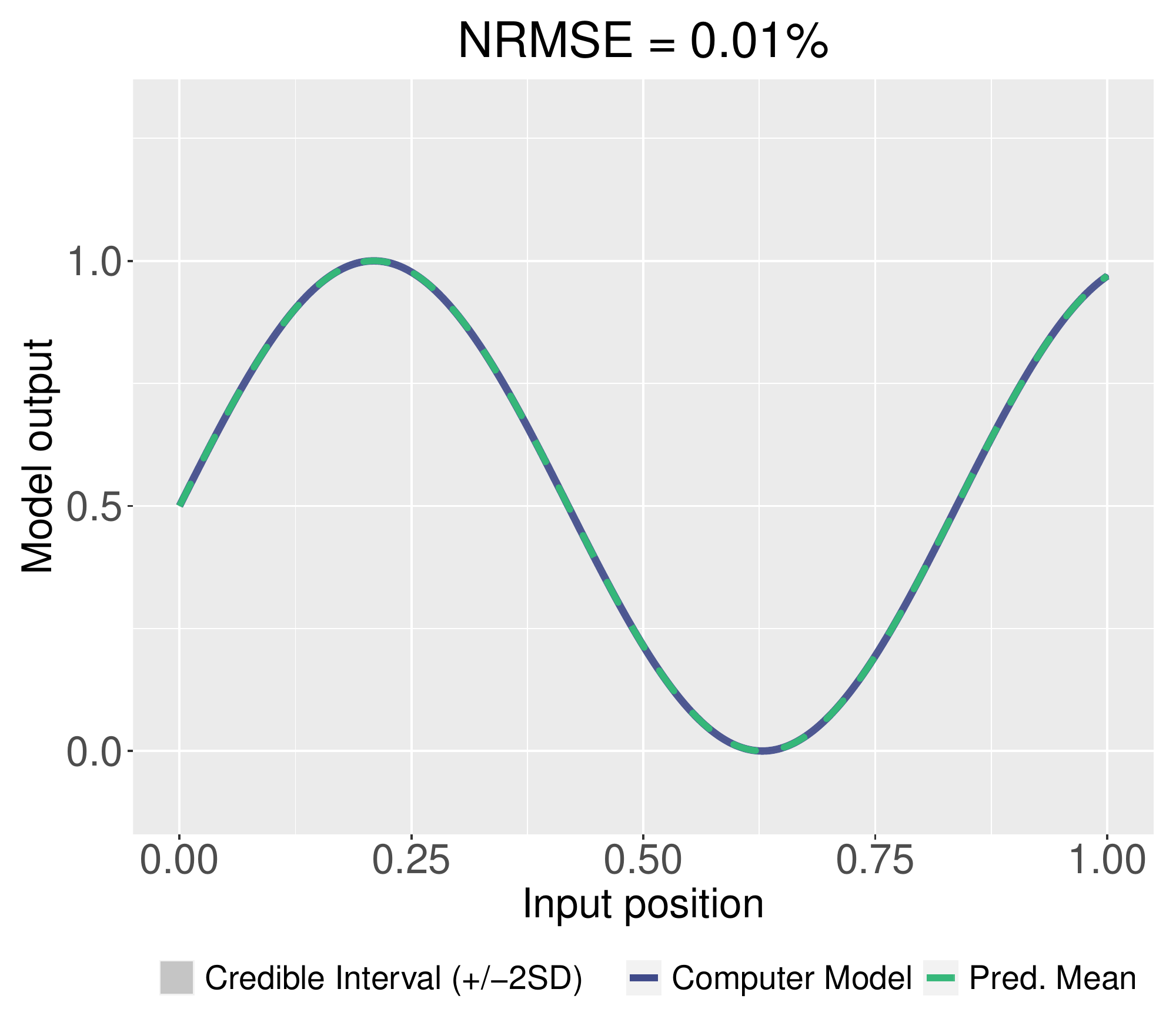}}
\subfloat[DGP Emulator of $f_2$]{\label{fig:dgp2}\includegraphics[width=0.33\linewidth]{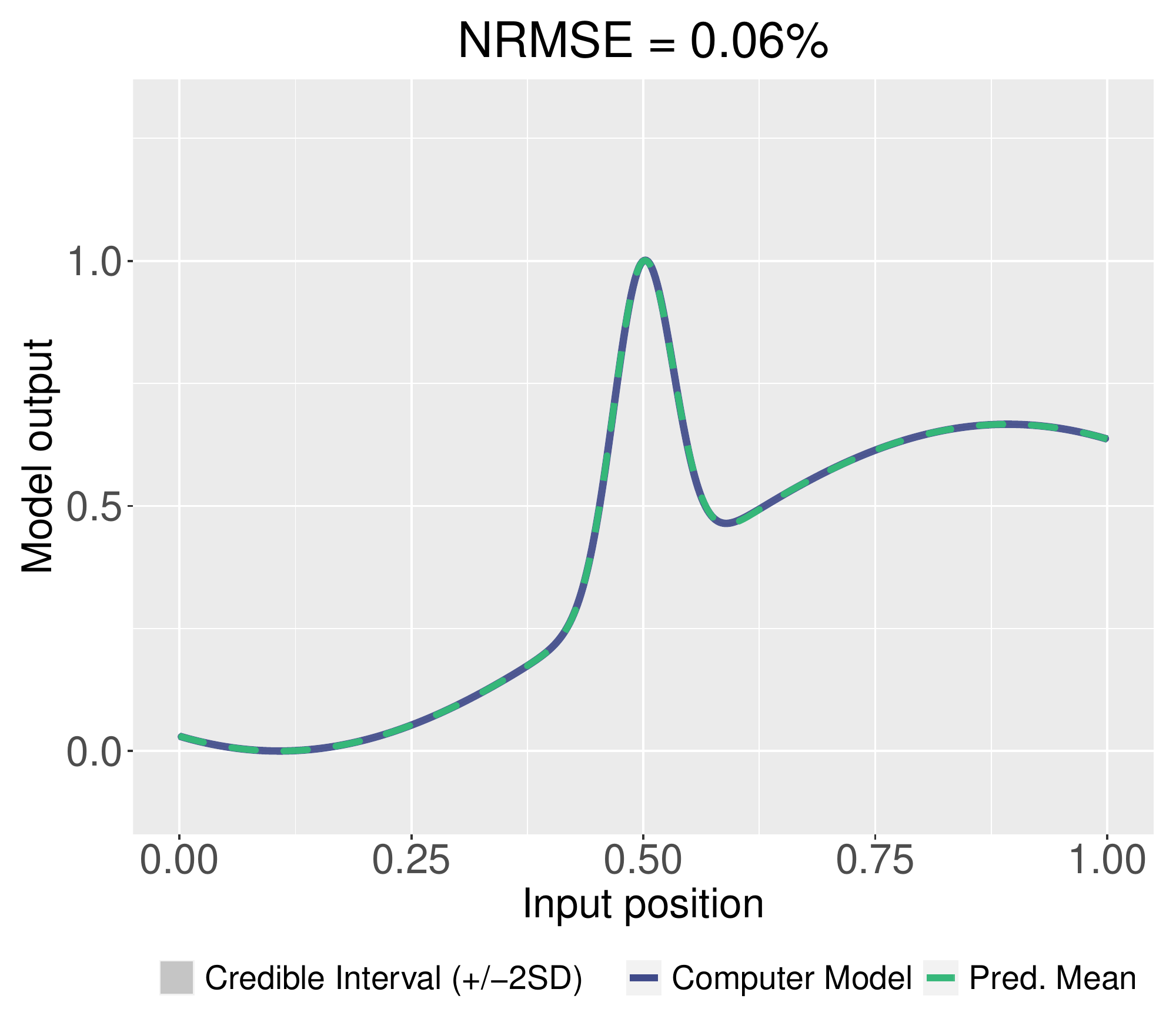}}
\subfloat[DGP Emulator of $f_3$]{\label{fig:dgp3}\includegraphics[width=0.33\linewidth]{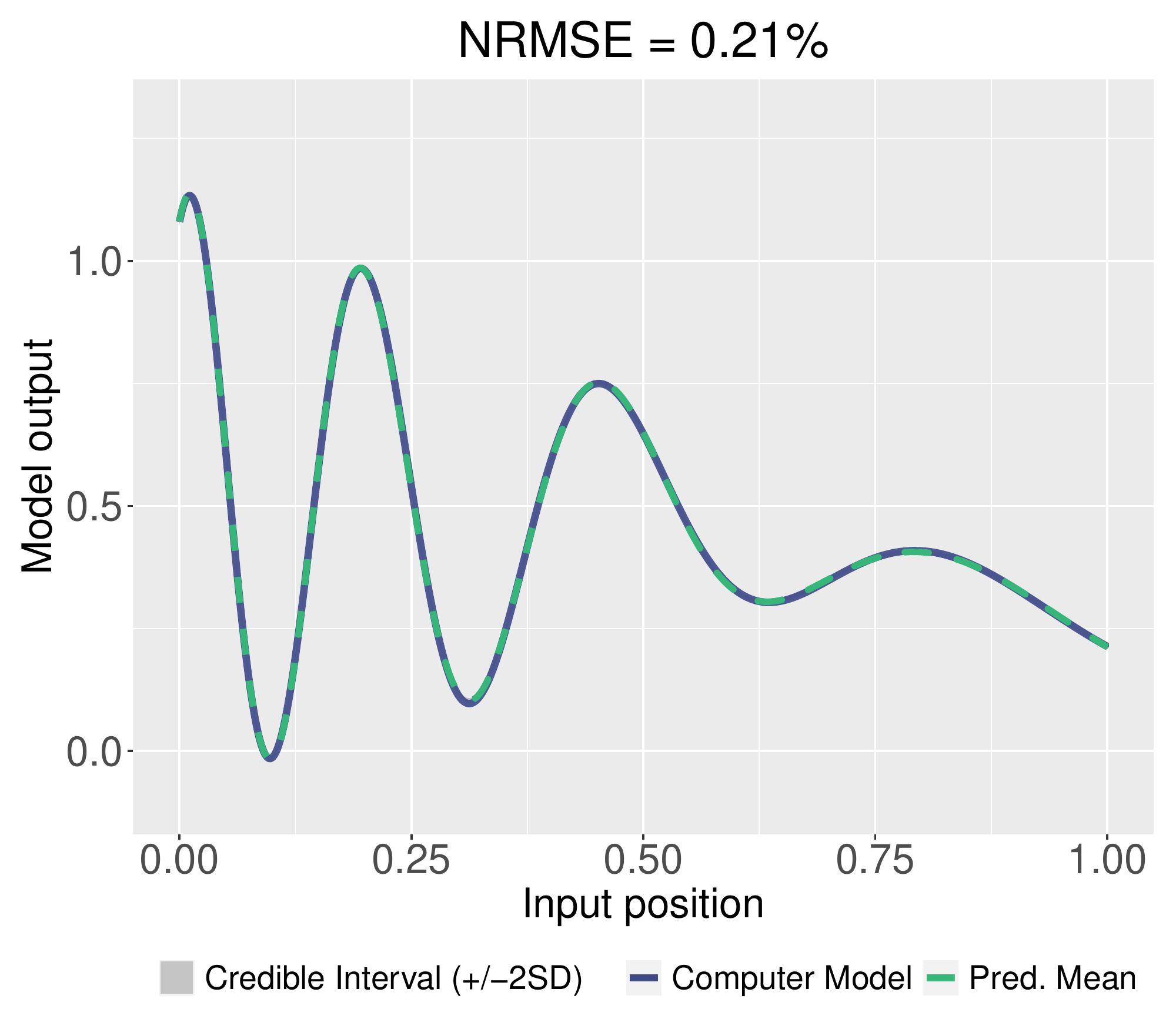}}
\caption{CGP emulator, CDGP emulator, LGP emulator (with its GP components), and LDGP emulator (with its DGP components) of the feed-forward synthetic network composed by the three computer models $f_1$, $f_2$ and $f_3$ specified in~\eqref{eq:f1},~\eqref{eq:f2} and~\eqref{eq:f3}. The NRMSEs are computed by normalizing the RMSEs with the range of the outputs in the testing dataset. The testing dataset is obtained by evaluating the synthetic network or the individual computer models at $500$ input positions uniformly spaced over $[0,1]$.}
\label{fig:1d}
\end{figure}

The constructed CGP emulator, CDGP emulator, LGP emulator (with its GP components), and LDGP emulator (with its DGP components) are shown in Figure~\ref{fig:1d}. The LGP and LDGP emulators both outperform the CGP and CDGP emulators and are able to mimic the rough fluctuations of the underlying network over $[0.4,0.5]$ and $[0.75,0.9]$ by exploiting the structural information of the network. By capturing the non-stationarity that exists in $f_2$ and $f_3$ through DGP emulators, the LDGP emulator achieves a significantly higher accuracy than the LGP emulator. As a result, the LDGP emulator offers the best predictive performance with only $0.18\%$ Normalized Root Mean Squared Error (NRMSE) given 20 evaluations of the network.

\section{Delta-Vega Hedging}
\label{sec:empirical}
Hedging is a crucial trading strategy utilized by financial engineers to minimize the risk associated with their investment portfolios. This is achieved by taking opposite positions in assets, such as options and their underlying stocks, within a portfolio. Option Greeks, such as Delta and Vega, are quantities that measure the sensitivities of an option with respect to features such as the spot price and volatility of underlying assets. They are important ingredients commonly used to establish risk hedging strategies. A hedging strategy typically uses Greeks of assets contained in a portfolio to optimize the asset allocations by neutralizing the risk (i.e., the overall Greeks) of the portfolio. Greek calculations can be computationally expensive for some assets, e.g., exotic options, and hedging strategies are often black boxes and can involve complex optimizations, especially when a portfolio contains a tremendous number of assets~\citep{horasanli2008hedging}. To ensure that the risk of a portfolio is hedged constantly and dynamically, we can emulate the hedging decision network, a chain of Greek calculations and portfolio optimizations, to render real-time rebalancing of asset positions within the portfolio.  

\begin{figure}[htbp]
\centering
\begin{tikzpicture}[shorten >=1pt,->,draw=black!50, node distance=4cm]
    \tikzstyle{every pin edge}=[<-,shorten <=1pt]
    \tikzstyle{neuron}=[circle,fill=vir2,minimum size=27.5pt,inner sep=0pt]
    \tikzstyle{layer1}=[neuron, fill=vir1];
    \tikzstyle{layer2}=[neuron, fill=vir3];
    \tikzstyle{layer3}=[neuron, fill=vir2];
    \tikzstyle{annot} = [text width=2.5em, text centered]

    \node[layer3,pin={[pin edge={<-}]left: $K_1,\,\tau_1,\,S_t$}] (I-1) at (0,0) {$\mathcal{V}$};
    \node[layer3,pin={[pin edge={<-}]left: $K_2,\,\tau_2,\,S_t$}] (I-2) at (0,-1.2) {$\mathcal{V}$};
    \node[layer2] (K-1) at (2.5, -0.6) {$\mathcal{H}_{\mathcal{V}}$};
    \node[layer2,pin={[pin edge={->}]right: $P_s$}] (J-1) at (5, -0.6) {$\mathcal{H}_{\Delta}$};
    \node[layer3,pin={[pin edge={<-}]left: $K_1,\,\tau_1,\,S_t$}] (L-1) at (4.1,-2.5) {$\Delta$};
    \node[layer3,pin={[pin edge={<-}]right: $K_2,\,\tau_2,\,S_t$}] (L-2) at (5.9,-2.5) {$\Delta$};
    \draw[->] (I-1) -- (K-1) node[midway,fill=white] {$\mathcal{V}_1$};
    \draw[->] (I-2) -- (K-1) node[midway,fill=white] {$\mathcal{V}_2$};
    \draw[->] (K-1) -- (J-1) node[midway,fill=white] {$P_2$};
    \draw[->] (L-1) -- (J-1) node[midway,fill=white] {$\Delta_1$};
    \draw[->] (L-2) -- (J-1) node[midway,fill=white] {$\Delta_2$};
\end{tikzpicture}
\caption{The hedging decision network that uses the Delta-Vega hedging strategy to neutralize the risk of a portfolio formed by two European call options (with strike prices $K_{i=1,2}$ and time-to-maturities $\tau_{i=1,2}$) and their underlying stock (with the spot price $S_t$). $\Delta$ and $\mathcal{V}$ represent respectively calculators that compute the Delta $\Delta_{i=1,2}$ and Vega $\mathcal{V}_{i=1,2}$ values of the two European call options. $P_2$ and $P_s$ are required positions of the second European call option and the underlying stock computed by the Vega strategy ($\mathcal{H}_{\mathcal{V}}$) and Delta strategy ($\mathcal{H}_{\Delta}$) such that the Delta and Vega values of the portfolio are minimized, given an unit position of the first European call option.}
\label{fig:delta-vega}
\end{figure}
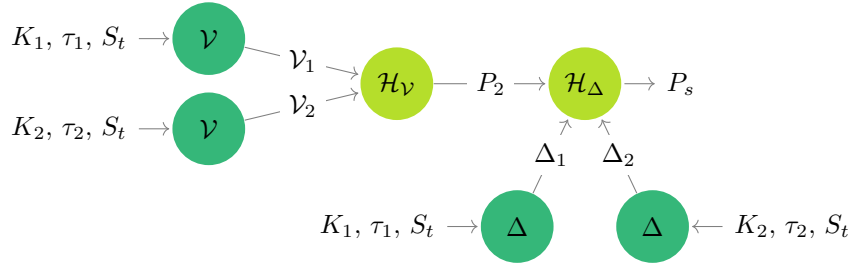

We consider a hedging decision network (shown in Figure~\ref{fig:delta-vega}) that is built on the Delta-Vega hedging strategy. This network is applied on a portfolio comprising two European call options (with values $C_{t,1}$ and $C_{t,2}$) and their underlying stock (with volatility, $v$, and spot price, $S_t$) at time $t$. The positions of the two options are denoted by $P_1$ and $P_2$, while the stock position is denoted by $P_s$. The network is a composite of four models. The first pair, the Vega ($\mathcal{V}$) and Delta ($\Delta$) calculators, calculate the Vega and Delta values, denoted by $\mathcal{V}_i=\partial C_{t,i}/\partial v$ and $\Delta_i=\partial C_{t,i}/\partial S_t$ respectively for $i=1,2$, per unit position of the two European call options. The input to $\mathcal{V}$ and $\Delta$ are options' strike prices ($K_1$ and $K_2$), time-to-maturities ($\tau_1$ and $\tau_2$ in years), and $S_t$. The third model in the network is the Vega strategy ($\mathcal{H}_{\mathcal{V}}$) that takes $\mathcal{V}_1$ and $\mathcal{V}_2$ as input and computes $P_2$ such that the Vega value of the portfolio is minimized, given a unit position of the first call option (i.e., $P_1=1$). Note that $\mathcal{H}_{\mathcal{V}}$ does not take the underlying stock into account as its Vega is zero. The final model of the network is the Delta strategy ($\mathcal{H}_{\Delta}$) that computes $P_s$ to neutralize the portfolio's Delta value given a unit position of $\Delta_1$ from the first option and $P_2$ positions of $\Delta_2$ from the second option. In the remainder of the section, we consider $\mathcal{V}$ and $\Delta$ that are derived from the Black-Scholes model~\citep{black1973pricing} governed by the Black-Scholes partial differential equation (PDE):
\begin{equation*}
    \frac{\partial C_t}{\partial t} + \frac{1}{2} v^2 S_t^2 \frac{\partial^2 C_t}{\partial S_t^2} + rS_t\frac{\partial C_t}{\partial S_t} - rC_t = 0
\end{equation*}
with the terminal condition $C_T=\max(S_T-K,0)$, where $C_t$ is the price of an European call option at time $t$, $K$ is the strike price of the option, $T=t+\tau$ is the option's maturity, $r$ is the risk-free interest rate, and $v$ is the stock volatility. 

To emulate the hedging decision network with respect to $P_2$ and $P_s$, we consider four emulator candidates, namely CGP, CDGP, LGP, and LDGP emulators. For CGP and CDGP, we focus on the global input $(S_t,\,K_1,\,K_2,\,\tau_1,\,\tau_2)$ over $[50,150]^3\times[1,2]^2$. For LGP and LDGP, we focus on the input $(S_t,\,K,\,\tau)$ over $[50,150]^2\times[1,2]$ for both $\mathcal{V}$ and $\Delta$, $(\mathcal{V}_1,\,\mathcal{V}_2)$ over $[0,78]^2$ for $\mathcal{H}_{\mathcal{V}}$, and $(\Delta_1,\,\Delta_2,\,P_2)$ over $[0,1]^2\times[0,100]$ for $\mathcal{H}_{\Delta}$. For all candidates, $r$ and $v$ in the Black-Scholes PDE are fixed to $0.05$ and $0.32$ respectively. For CGP and CDGP emulators, we initialize the training dataset with a design size of 50 over $(S_t,\,K_1,\,K_2,\,\tau_1,\,\tau_2)$ that is drawn from the maximin Latin Hypercube sampler. For LGP and LDGP emulators, we build GP and DGP emulators of $\mathcal{V}$, $\Delta$, $\mathcal{H}_{\mathcal{V}}$ and $\mathcal{H}_{\Delta}$ independently with initial maximin Latin Hypercube design sizes of 30, 30, 20 and 30 respectively. We then refine the CGP emulator, CDGP emulator, and individual GP and DGP emulators within LGP and LDGP emulators by enriching their designs sequentially until the training datasets reach $100$ design points. For each emulator, we repeat the training $30$ times and choose the emulator with the lowest RMSEs using $500$ validation points for the final comparison. 

\begin{figure}[ht!]
\centering 
\includegraphics[width=0.75\linewidth]{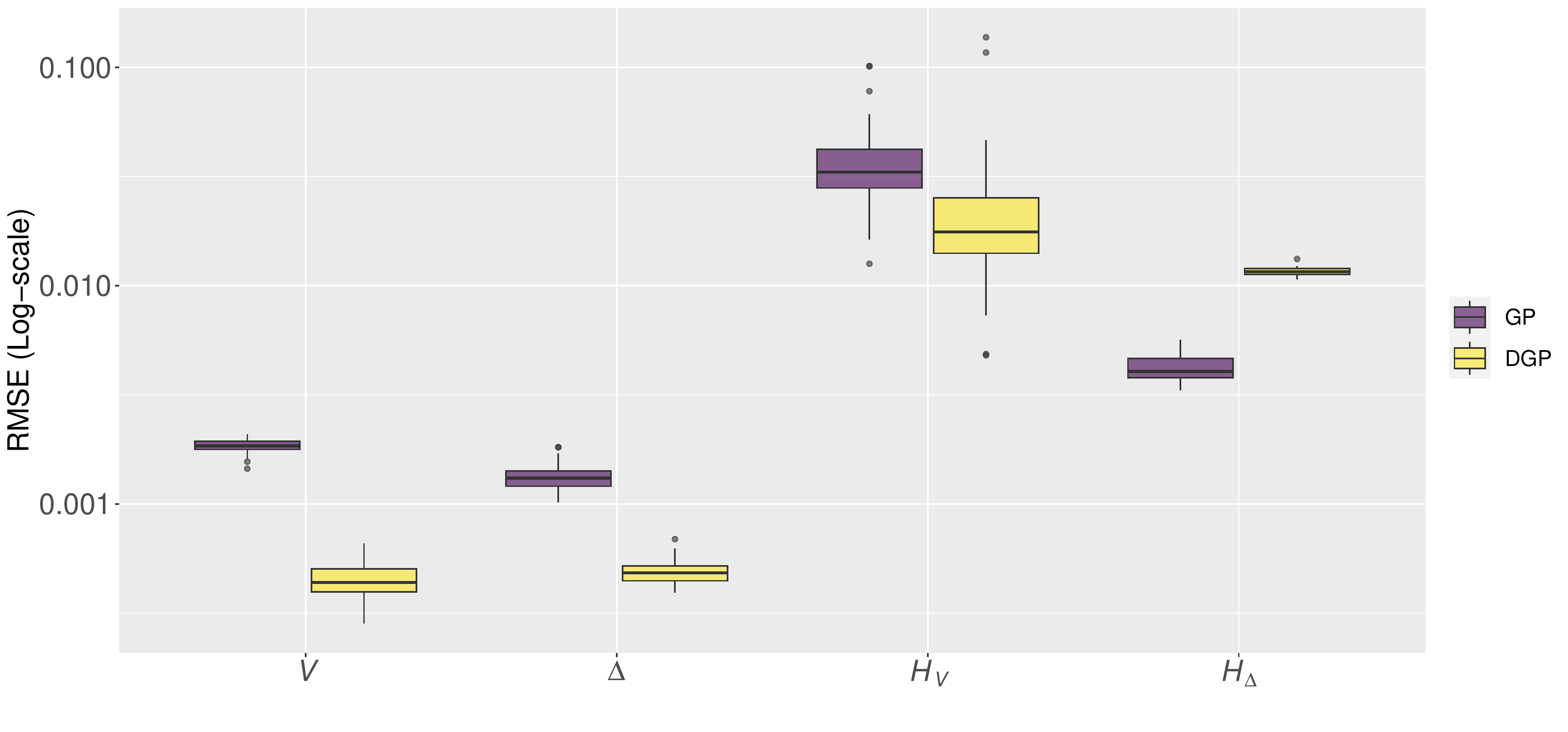}
\caption{RMSEs of the GP and DGP emulators of $\mathcal{V}$, $\Delta$, $\mathcal{H}_{\mathcal{V}}$, and $\mathcal{H}_{\Delta}$ that are evaluated at $500$ validation data points across $30$ training trials.}
\label{fig:comparison}
\end{figure}

Figure~\ref{fig:comparison} presents the RMSEs of the individual GP and DGP emulators for $\mathcal{V}$, $\Delta$, $\mathcal{H}_{\mathcal{V}}$, and $\mathcal{H}_{\Delta}$. Whilst the DGP emulators were superior to GP for $\mathcal{V}$, $\Delta$ and $\mathcal{H}_{\mathcal{V}}$, they overfit for $\mathcal{H}_{\Delta}$ as GP emulators already reached sufficiently low normalized RMSEs (averaged around $0.0044\%$). We therefore construct our final LDGP emulator by linking the best (in terms of RMSE) DGP emulators of $\mathcal{V}$, $\Delta$, and $\mathcal{H}_{\mathcal{V}}$ with the best GP emulator of $\mathcal{H}_{\Delta}$.

\begin{figure}[ht!]
\centering 
\subfloat[CGP ($P_2$)]{\label{fig:cgp_option}\includegraphics[width=0.25\linewidth]{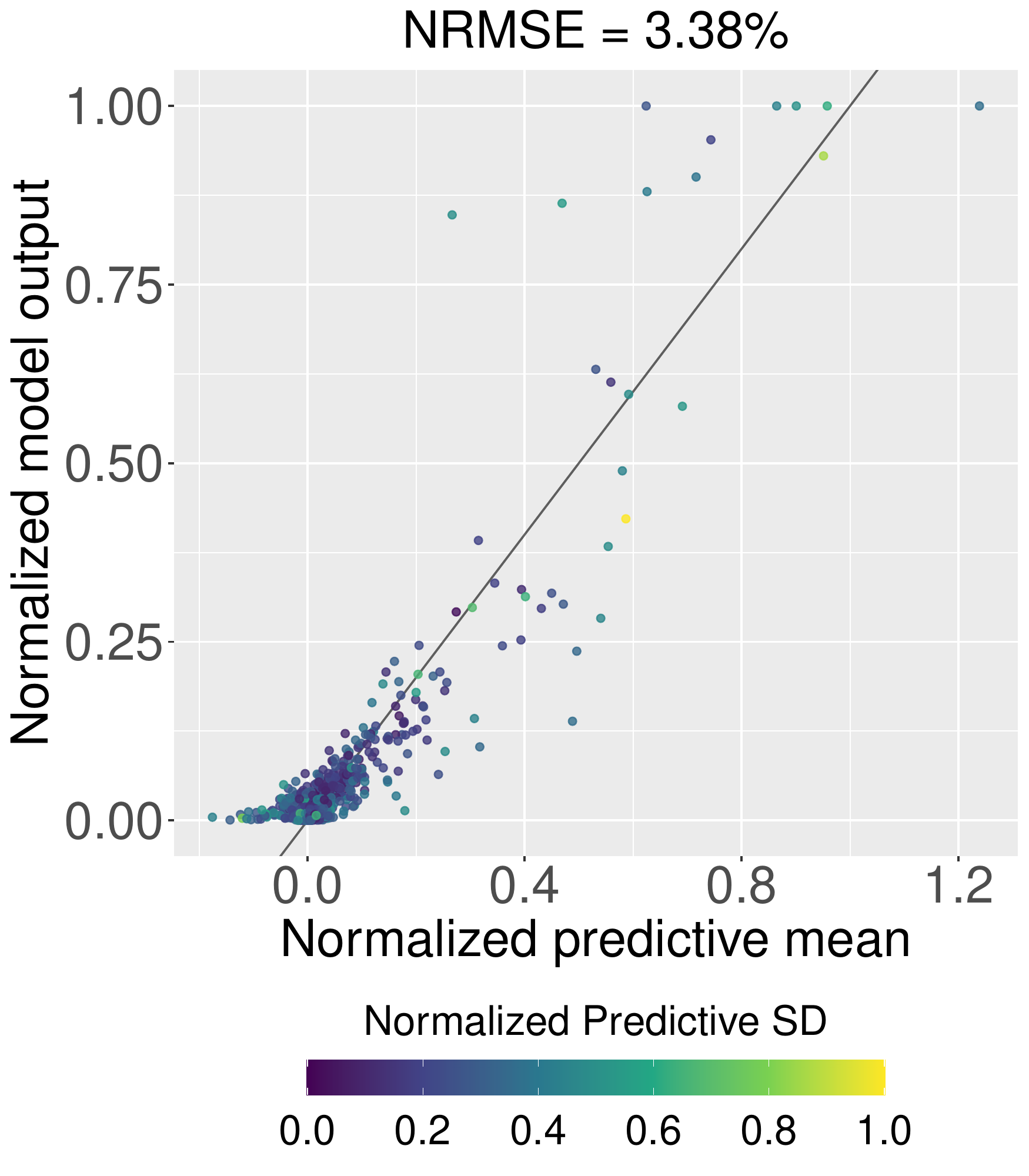}}
\subfloat[CDGP ($P_2$)]{\label{fig:cdgp_option}\includegraphics[width=0.25\linewidth]{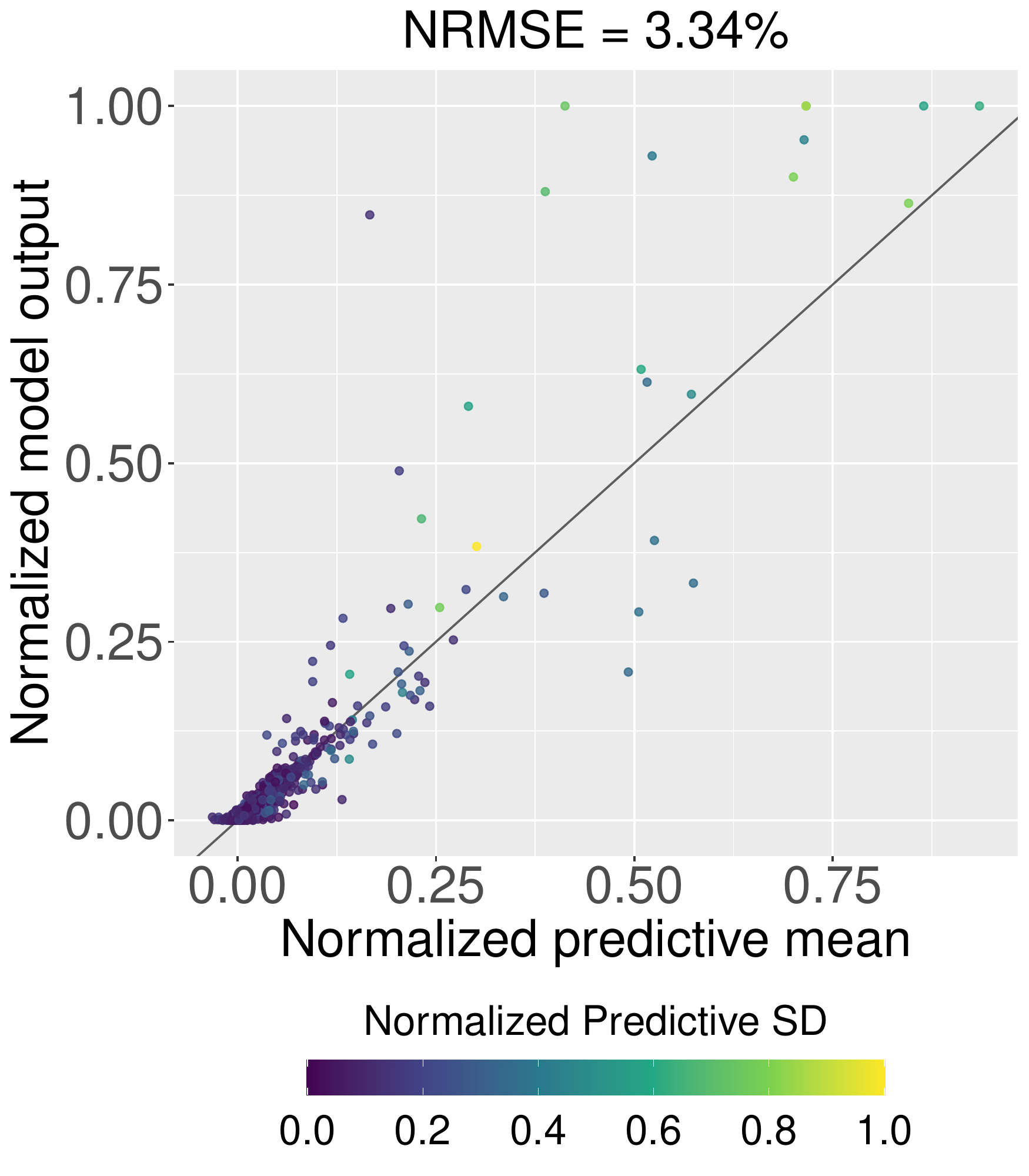}}
\subfloat[LGP ($P_2$)]{\label{fig:lgp_option}\includegraphics[width=0.25\linewidth]{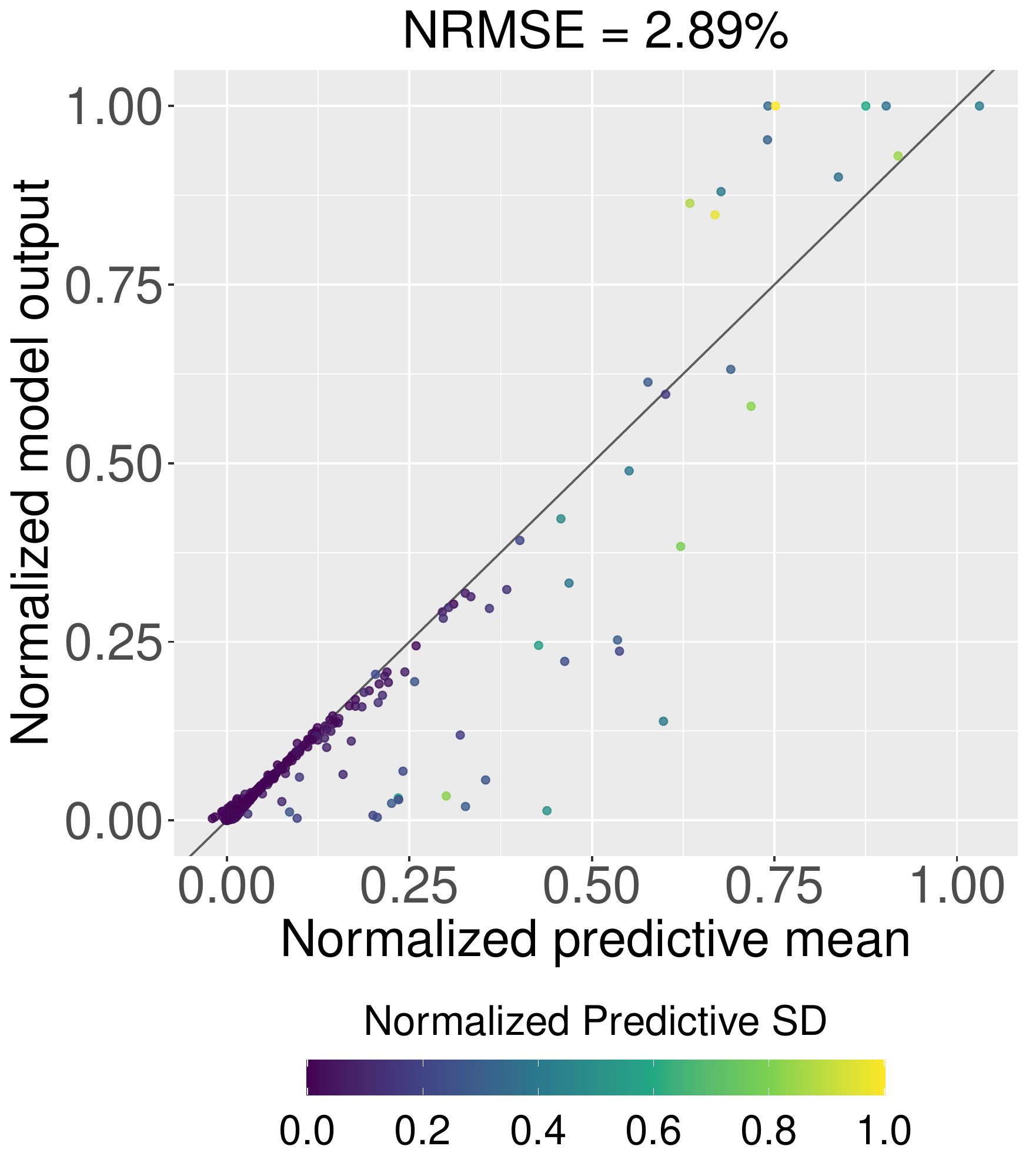}}
\subfloat[LDGP ($P_2$)]{\label{fig:ldgp_option}\includegraphics[width=0.25\linewidth]{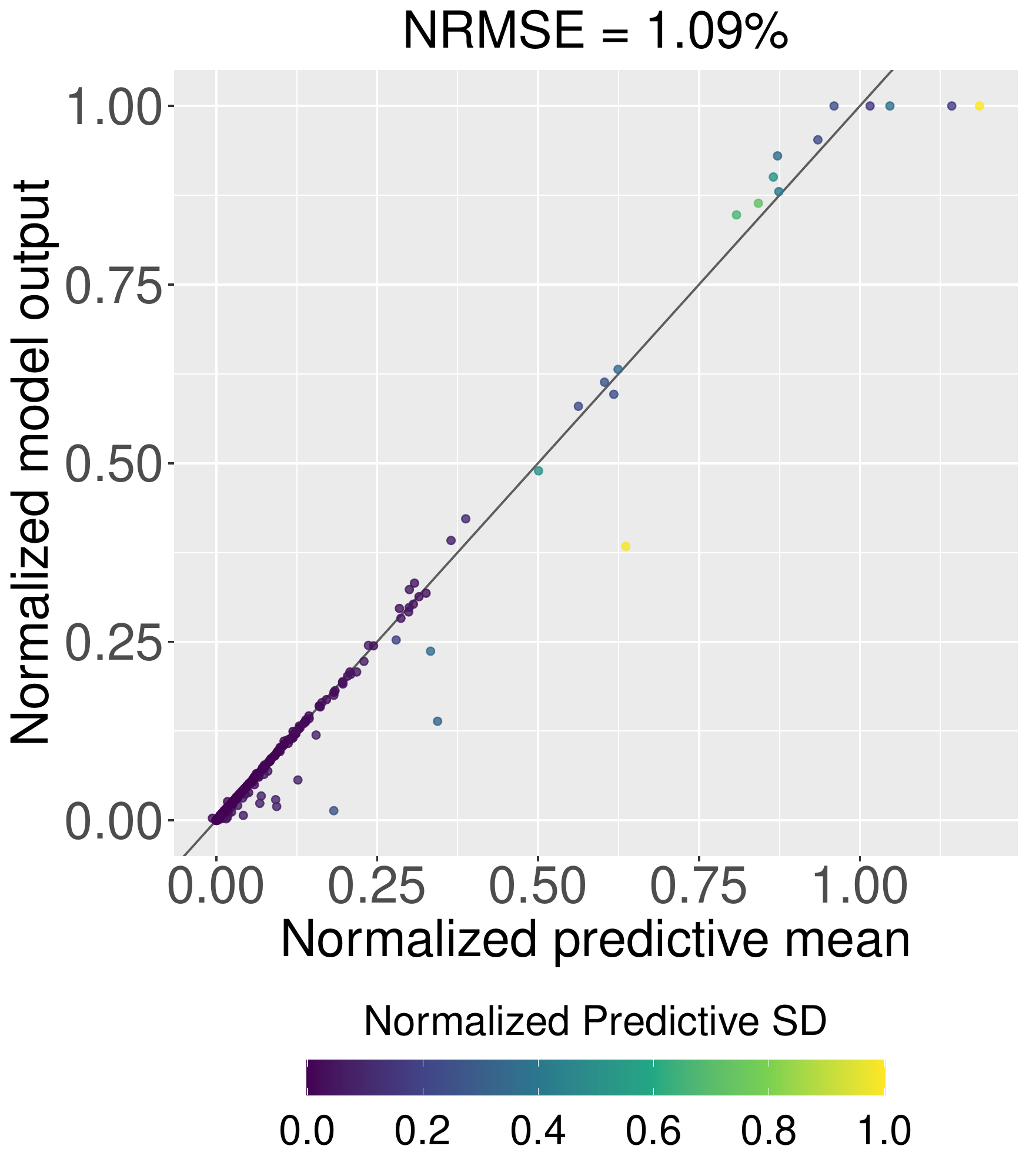}}\\
\subfloat[CGP ($P_s$)]{\label{fig:cgp_stock}\includegraphics[width=0.25\linewidth]{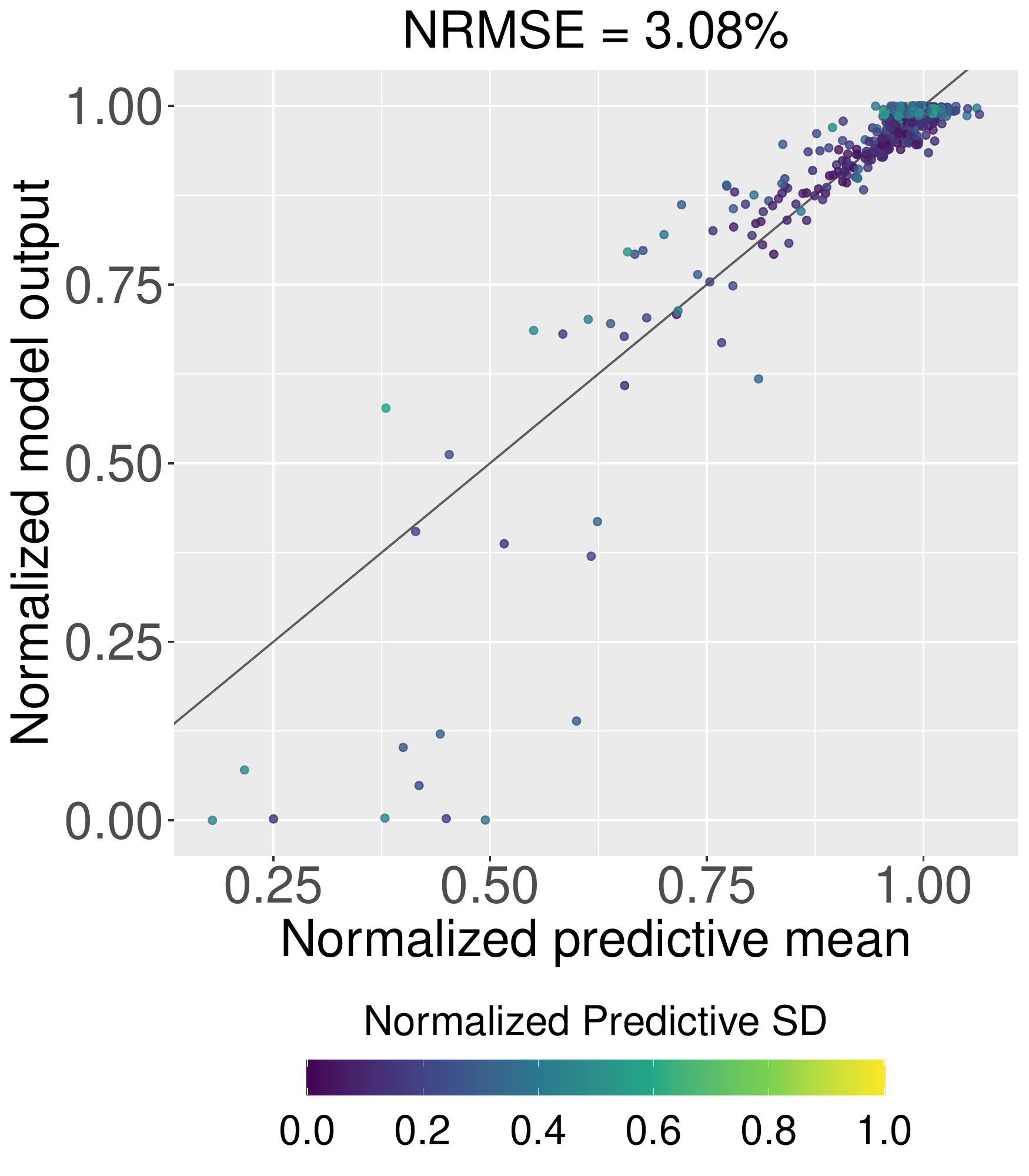}}
\subfloat[CDGP ($P_s$)]{\label{fig:cdgp_stock}\includegraphics[width=0.25\linewidth]{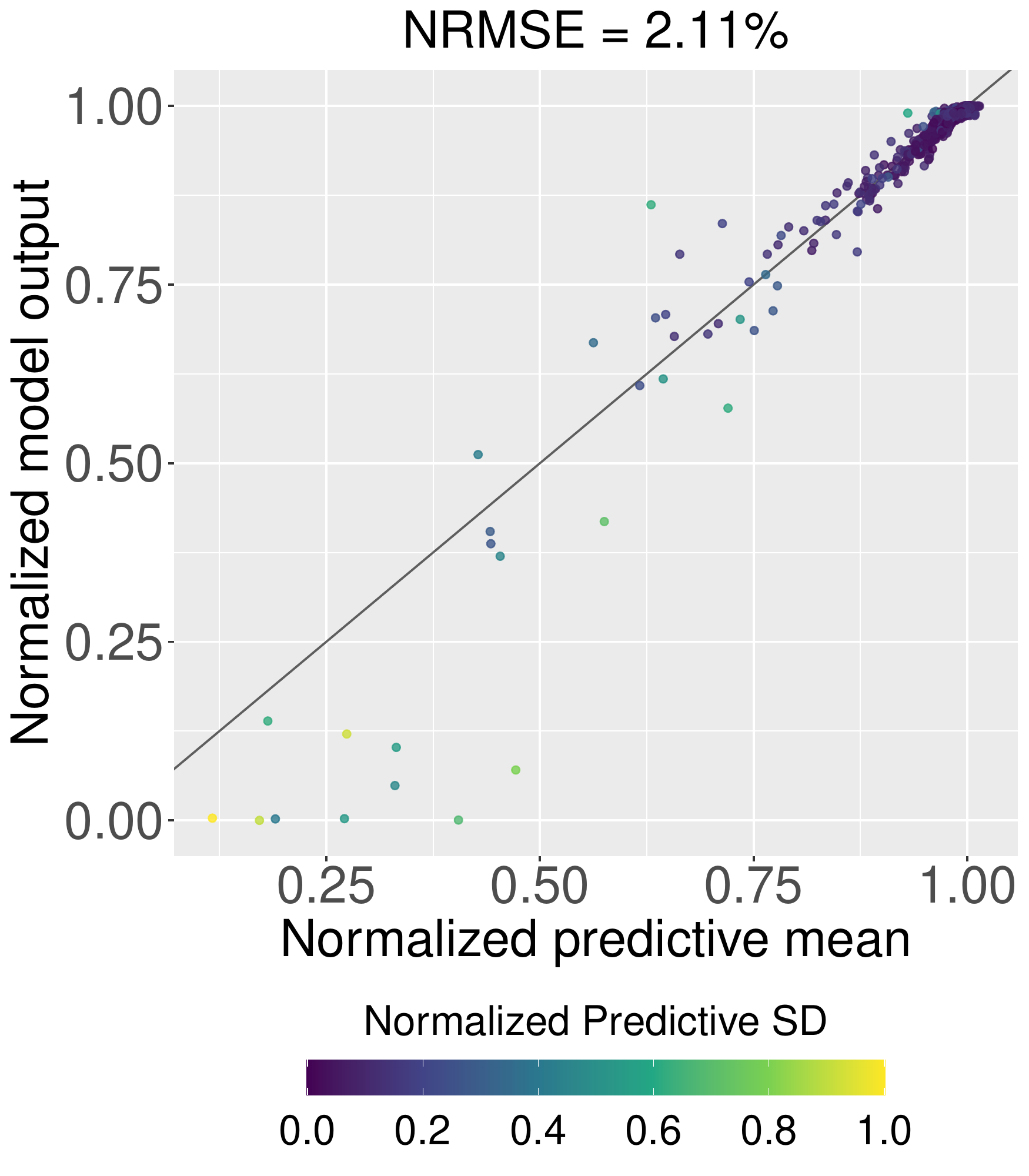}}
\subfloat[LGP ($P_s$)]{\label{fig:lgp_stock}\includegraphics[width=0.25\linewidth]{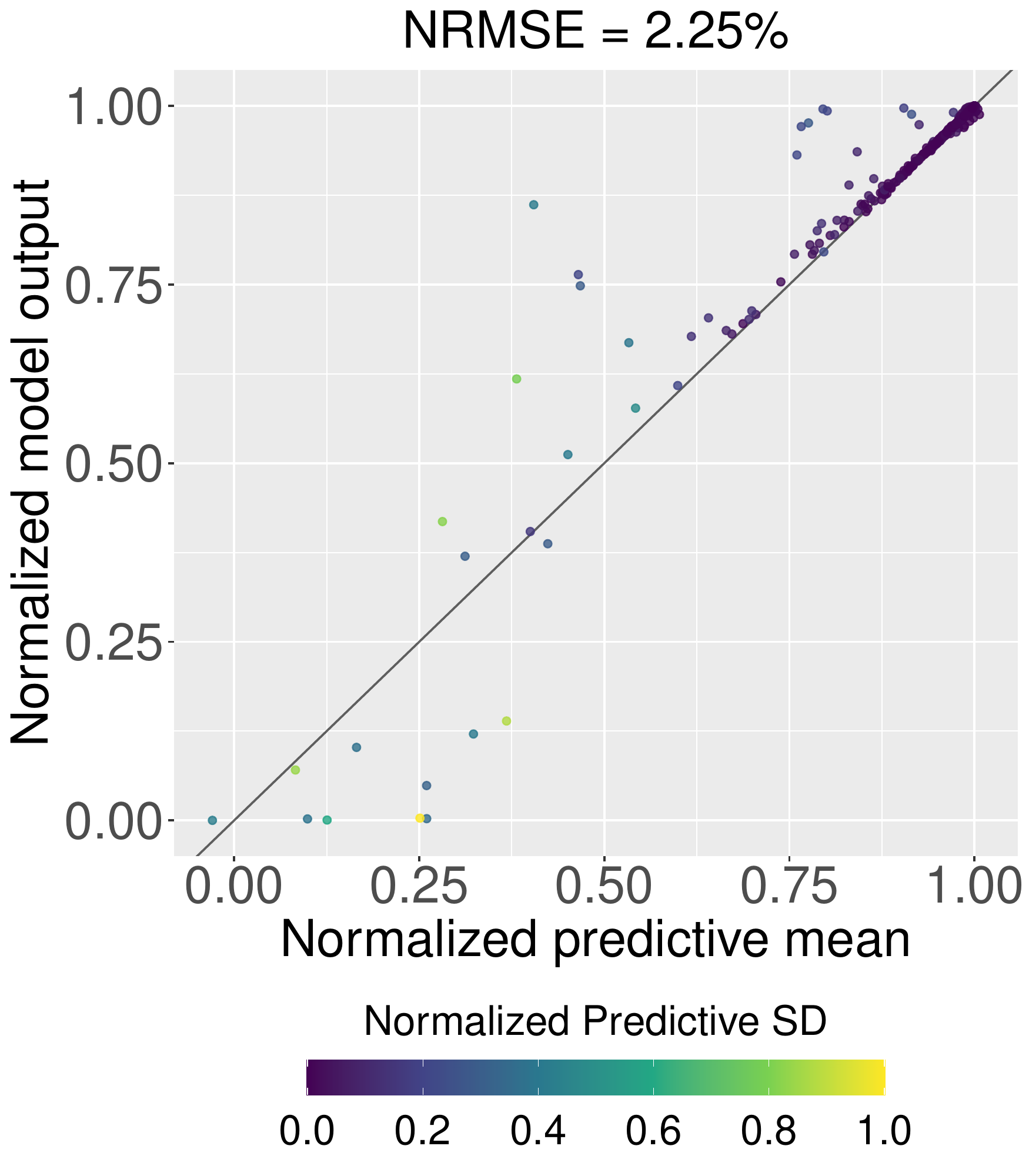}}
\subfloat[LDGP ($P_s$)]{\label{fig:ldgp_stock}\includegraphics[width=0.25\linewidth]{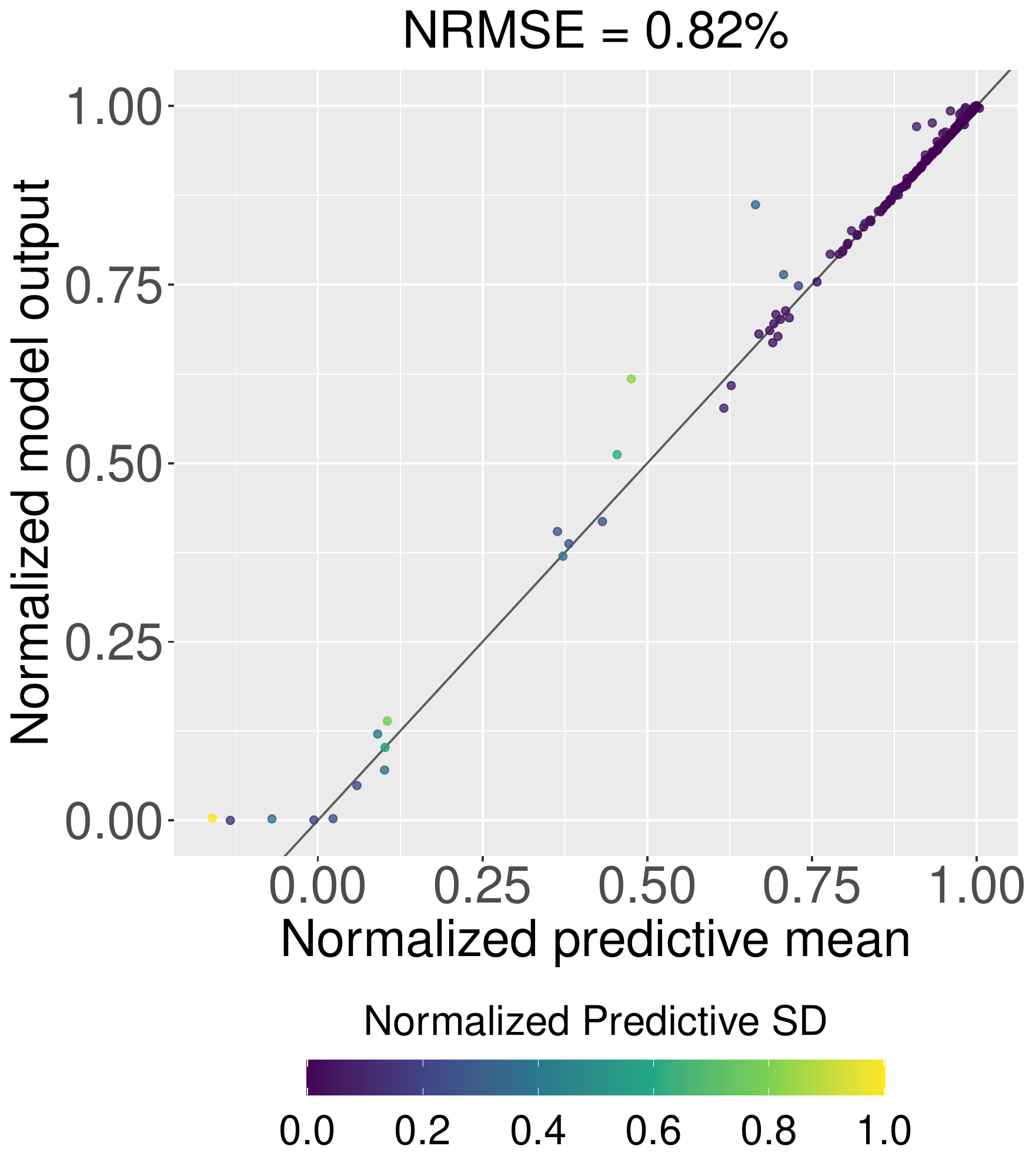}}
\caption{Plots of $P_2$ and $P_s$ (normalized by their max and min values) from the hedging decision network~\ref{fig:delta-vega} at $2000$ testing input positions \emph{vs} the min-max normalized mean predictions, along with min-max normalized predictive standard deviations, made by the CGP, CDGP, LGP, and LDGP emulators trained with $100$ design points and squared exponential kernels.}
\label{fig:hedging}
\end{figure}

Figure~\ref{fig:hedging} compares the emulation performance of the final CGP, CDGP, LGP and LDGP emulators with respect to the global outputs, $P_2$ and $P_s$, of the hedging decision network using $2000$ testing data points. It can be seen from Figure~\ref{fig:hedging} that CDGP emulators perform better than the CGP emulators since the CDGP emulators capture the non-stationary behaviors exhibited in the network. With regard to the CDGP and LGP emulators, though the LGP emulator provides lower NRMSE for $P_2$ it gives a slightly higher NRMSE for $P_s$. This observation can be explained by noting that \begin{enumerate*}[label=(\roman*)]
  \item the LGP emulator can be seen as a DGP emulator with latent layers fully exposed and thus is less flexible representing the complex behaviors, e.g., the non-stationarity of $\mathcal{V}$ and $\Delta$~\citep{ming2022deep}, embedded in the network;
  \item the DGP emulator does not incorporate the structural information of the network and thus can result in inferior performance than the LGP emulator, which effectively reduces the input dimension by decoupling the network into simpler emulation problems.
\end{enumerate*}
The trade-off of effects from these two aspects thus can lead to the different performance of CDGP and LGP emulators with respect to $P_2$ and $P_s$. Since LDGP emulators exploit both the structure of the hedging decision network and the non-stationarity embedded in the network, they deliver significantly better predictive accuracy and uncertainties than the CDGP and LGP emulators.   

It's also worth noting that both LGP and LDGP emulators are less computationally expensive to construct compared to the CGP and CDGP emulators. This is because both CGP and CDGP emulators require generations of design points by invoking both $\mathcal{V}$ and $\Delta$ twice, while LGP and LDGP emulators only require invoking $\mathcal{V}$ and $\Delta$ once, resulting in reduced computational burden during construction. Furthermore, constructing CGP and CDGP emulators requires building two separate emulators for $P_2$ and $P_s$ independently, whereas only a single LGP and LDGP emulator is required for both outputs.

\section{Discussion}
\label{sec:conclusion}
The propagation of uncertainty through chains or networks of models from different disciplines is an important challenge within the UQ community. In this work we represent a network of computer models as a deep Gaussian process with partial exposure of its hidden layers. We adapt the Stochastic Imputation algorithm for inference on DGPs~\citep{ming2022deep} to accommodate the partial exposure, resulting in the Linked Deep Gaussian Process (LDGP). We demonstrated the efficacy of LDGPs in comparison with LGPs (which require stationary GP emulator representations for each computer model in the network) and both the composite DGP and stationary GP emulators (where the network is effectively treated as a black box with its output emulated as a function of its input). 

In all of our test cases the LDGP emulator considerably outperforms the other emulators. The reason for this out-performance lies broadly in the fact that the LDGP, by accommodating the output from the component models of the network, is able to account for more data. Moreover, by constructing component DGPs individually, our sequential design strategy is able to ensure these components are emulated well over the range of possible inputs (rather than this range being artificially limited by the design on the overall network inputs).

As with LGP emulation, LDGP emulation is best suited in scenarios where a network of computer models can be broken down into sub-components with relatively simple functional behaviors. If the internal models exhibit particularly difficult (to emulate) behavior, the uncertainty in that component DGP will propagate through the LDGP and may compromise performance compared to a composite DGP. A promising avenue for future research would involve exploring methods to optimally decouple a computer model network for efficient uncertainty quantification.

The extension of LDGPs (and indeed LGPs) to networks of models with complex output structures represents an important challenge. Certain structures, such as the output of high-dimensional spatio-temporal fields, are ubiquitous in environmental decision making, where a land use change and changing climate might both feed into subsequent models for carbon sequestration, biodiversity, flood risk and so on. In these cases, projection of the high-dimensional field onto a low-dimensional basis, treating the basis coefficients as the model output (for emulation) and then, within the downstream model, projecting back into the spatio-temporal field required to drive it, may be a way to preserve a tractable DGP representation of the network, though projection can induce added uncertainties. The handling of internal models with non-continuous output types (e.g., counts or categories), within this or a similar framework requires further research.

\section*{Acknowledgments}
Deyu Ming and Daniel Williamson are funded by the BBSRC greenhouse gas removal demonstrator, Net Zero Plus, BB/V011588/1 and by an EPSRC AI for Net Zero Project, EP/Y005597/1.

\bibliographystyle{agsm}  
\bibliography{references}  

\begin{appendices}
\setcounter{equation}{0}
\setcounter{section}{0}
\setcounter{algorithm}{0}
\makeatletter
\renewcommand{\theequation}{A\arabic{equation}} 
\renewcommand{\thealgorithm}{A\arabic{algorithm}} 

\section{Elliptical Slice Sampling Algorithm}
\label{app:ess}
The Elliptical Slice Sampling (ESS) can be used to sample from the posterior $\pi(\mathbf{w})$ of the latent variable $\mathbf{w}\in\mathbb{R}^{M\times 1}$ in the following form
\begin{equation}
\label{eq:ess}
\pi(\mathbf{w})\propto\mathcal{L}(\mathbf{w})\mathcal{N}(\mathbf{w};\boldsymbol{\mu},\boldsymbol{\Sigma}),
\end{equation}
where $\mathcal{L}(\mathbf{w})$ is a likelihood function and $\mathcal{N}(\mathbf{w};\boldsymbol{\mu},\boldsymbol{\Sigma})$ is a multivariate normal prior of $\mathbf{w}$ with mean $\boldsymbol{\mu}$ and covariance matrix $\boldsymbol{\Sigma}$. Note that given $\{\mathbf{W}_{l-1,p}\}_{p=1,\dots,P_{l-1}}$ and $\{\mathbf{W}_{l+1,p}\}_{p=1,\dots,P_{l+1}}$, the conditional posterior of the latent output $\{\mathbf{W}_{l,p}\}_{p=1,\dots,P_l}$ from layer $l$ for all $l=1,\dots, L-1$ can be expressed in the form of~\eqref{eq:ess}:
\begin{multline}
\label{eq:layer_pos}
p(\{\mathbf{w}_{l,p}\}_{p=1,\dots,P_l}|\{\mathbf{w}_{l-1,p}\}_{p=1,\dots,P_{l-1}}, \{\mathbf{w}_{l+1,p}\}_{p=1,\dots,P_{l+1}})\propto\\ \prod_{p=1}^{P_{l+1}}p(\mathbf{w}_{l+1,p}|\{\mathbf{w}_{l,p}\}_{p=1,\dots,P_l})\,\prod_{p=1}^{P_{l}}p(\mathbf{w}_{l,p}|\{\mathbf{w}_{l-1,p}\}_{p=1,\dots,P_{l-1}}),
\end{multline}
where all terms are multivariate normal; $p(\mathbf{w}_{1,p}|\{\mathbf{w}_{0,p}\}_{p=1,\dots,P_{0}})=p(\mathbf{w}_{1,p}|\mathbf{x})$ when $l=1$ and $p(\mathbf{w}_{L,p}|\{\mathbf{w}_{L-1,p}\}_{p=1,\dots,P_{L-1}})=p(\mathbf{y}_{p}|\{\mathbf{w}_{L-1,p}\}_{p=1,\dots,P_{L-1}})$ when $l=L-1$. As the result, we can draw an imputation of $\{\mathbf{W}_{l,p}\}_{p=1,\dots,P_l}$ from its conditional posterior using a single ESS update presented in Algorithm~\ref{alg:ess}.

\begin{algorithm}[!ht]
\caption{Elliptical Slice Sampling Update for $\{\mathbf{W}_{l,p}\}_{p=1,\dots,P_l}$}
\label{alg:ess}
\begin{algorithmic}[1]
\REQUIRE{Current state $\{\mathbf{w}_{l,p}\}_{p=1,\dots,P_l}$.}
\ENSURE{New state $\{\mathbf{w}^\prime_{l,p}\}_{p=1,\dots,P_l}$.}
\STATE{Form a sample $\boldsymbol{\nu}=[\boldsymbol{\nu}_1,\dots,\boldsymbol{\nu}_{P_l}]$, where $\boldsymbol{\nu}_p$ is drawn from the multivariate normal distribution defined by $p(\mathbf{w}_{l,p}|\{\mathbf{w}_{l-1,p}\}_{p=1,\dots,P_{l-1}})$ for $p=1,\dots,P_l$.}
\STATE{Draw a sample $u$ from the uniform distribution $\mathcal{U}(0,1)$.}
\STATE{Set log-likelihood threshold $\log\mathcal{L}(\boldsymbol{\nu})=\sum^{P_{l+1}}_{p=1}\log p(\mathbf{w}_{l+1,p}|\boldsymbol{\nu})+\log u$.}
\STATE{Draw an initial proposal $\theta$ from the uniform distribution $\mathcal{U}(0,2\pi)$.}
\STATE{Define a bracket $[\theta_{\mathrm{min}},\theta_{\mathrm{max}}]=[\theta-2\pi,\theta]$.}
\STATE{Compute $\mathbf{w}^\prime_{l,p}=\mathbf{w}_{l,p}\cos\theta+\boldsymbol{\nu}_p\sin\theta$ for all $p=1,\dots,P_l$.}\label{alg:marker}
\IF{$\log\mathcal{L}(\{\mathbf{w}^\prime_{l,p}\}_{p=1,\dots,P_l})>\log\mathcal{L}(\boldsymbol{\nu})$} 
   \RETURN $\{\mathbf{w}^\prime_{l,p}\}_{p=1,\dots,P_l}$
\ELSE
   \IF{$\theta<0$}
     \STATE{$\theta_{\mathrm{min}}=\theta$}
   \ELSE
    \STATE{$\theta_{\mathrm{max}}=\theta$}
    \ENDIF
    \STATE{Draw a proposal $\theta$ from the uniform distribution $\mathcal{U}(\theta_{\mathrm{min}},\theta_{\mathrm{max}})$.}
    \STATE{{\bf{goto}} \ref{alg:marker}}
\ENDIF
\end{algorithmic} 
\end{algorithm}
\end{appendices}


\end{document}